\documentclass[twoside]{article}

\usepackage[accepted]{aistats2026}
\usepackage{amsmath,amsfonts,bm,xcolor,amsthm}
\usepackage{graphicx}
\usepackage{subcaption}
\definecolor{darkblue}{rgb}{0,0,.5}
\usepackage[hyphens]{url}
\usepackage[colorlinks=true,allcolors=darkblue]{hyperref}
\usepackage{booktabs}

\newtheorem{theorem}{Theorem}[section]
\newtheorem{corollary}{Corollary}[theorem]
\newtheorem{lemma}{Lemma}[theorem]

\newtheorem{assumption}{Assumption}[theorem]
\newtheorem{definition}{Definition}[theorem]
\usepackage[round]{natbib}

\begin{document}

%

%

\twocolumn[

\aistatstitle{Two Mathematical Models of Knowledge Distillation}

\aistatsauthor{ Audrey Xie \And Ludwig Schmidt \And  John Duchi }

\aistatsaddress{ Stanford University \And  Stanford University \And Stanford University } ]

\begin{abstract}
  Many hypotheses compete to explain the successes of knowledge
  distillation.
  To help address this, we propose and analyze a mathematical model of
  distillation, which suggests that distillation's performance comes not
  from obtaining better models but from easier to optimize landscapes.
  For generalized linear models trained with stochastic gradient descent, we
  prove that distillation fits performant student models asymptotically more
  quickly than non-distilled models.
  In rank-1 matrix approximation, we
  characterize conditions on the target matrix under which gradient descent
  with distillation converges strictly faster than training on the
  supervised objective.
  The theory helps delineate the ways distillation
  provides benefits (i.e., in optimization speed, not
  in generalization), and experiments on real datasets
  corroborate the theoretical predictions.
\end{abstract}

\section{INTRODUCTION}\label{section:introduction}
Knowledge distillation transfers the predictive ability of a larger teacher model to a smaller student model \citep{BC14,HVD15} by training the student on teacher predictions rather than ground-truth labels.
It provides an important tool for obtaining performant models with reduced inference costs~\citep{RBKC14,KR16,SDCW19,BZRMA022,gemma24,SMJC24}.
Two perspectives compete to explain distillation's effectiveness: one
attributes its benefits to the student's improved generalization ability
\citep{MFB20,MRRKK20,DKSM21,DS23,PDO24,IGTMO25}; the other emphasizes its
algorithmic effects during optimization~\citep{PL19,SPA23,PLMRG25}.

Recent evidence suggests that there is still a gap in our understanding 
of knowledge distillation.
In particular, \citet{BSWRLW25} show empirically that, given sufficient
data, a student distilled from an optimally chosen teacher does not
outperform the same model trained directly on ground-truth labels. This
raises a fundamental question: if distillation does not improve the
asymptotic solution, why does it often accelerate learning in practice?

In this work, we provide a theoretical explanation by analyzing knowledge
distillation in two canonical settings: generalized linear models (GLMs) fit
with maximum likelihood via stochastic gradient descent and
rank-1 matrix approximation.
Our analysis shows that although a student
trained with distillation ultimately converges to the same solution as one
trained with ground-truth supervision, distillation improves the efficiency
of gradient descent algorithms. We test these theoretical predictions
through experiments on synthetic data, regression tasks, and language model
pretraining.
Our contributions follow:
\begin{itemize}
\item \textbf{Online SGD for GLMs.} We model the teacher as the population
  risk minimizer over the full covariates and the student as having fewer
  parameters.
  We prove that distillation and direct supervision converge to the same
  solution (Theorem~\ref{theorem:glm-as-convergence-pop}), but that
  distillation yields asymptotically more efficient SGD iterates, with the
  efficiency gap proportional to the label variance
  (Theorem~\ref{theorem:glm-asymp-normality-pop}).
\item \textbf{Finite-dataset SGD.} Extending these results, we show that
  both trajectories converge to the same point on finite datasets
  (Theorem~\ref{theorem:glm-as-convergence-emp}), and that, with high
  probability, distillation iterates have smaller asymptotic covariance
  (Theorem~\ref{theorem:glm-asym-normality-emp}).
\item \textbf{Rank-1 matrix approximation.} We model the teacher as the best
  rank-$r$ Frobenius approximation to a target matrix.
  We prove that gradient descent applied to either the target or the teacher
  approximation converges to the same set of global optima
  (Theorem~\ref{theorem:rank-1-convergence-to-mins}).
  Yet distillation converges faster under the sufficient condition that the
  matrix dimension scales inversely with the target error tolerance, raised
  to a spectrum-dependent constant
  (Theorem~\ref{theorem:conditions-for-kd-gd-faster}).
\item \textbf{Experiments.} We test our theoretical predictions on synthetic
  data, regression with Fashion-MNIST~\citep{XRV17}, and
  GPT-2 trained on C4~\citep{RWCLAS19, DKSM21}.
\end{itemize}

\section{PRELIMINARIES}\label{section:background}

We review knowledge distillation and formalize it for maximum likelihood
estimation in generalized linear models and low-rank matrix approximation.

\subsection{Knowledge Distillation}

Knowledge distillation trains student models with teacher-generated labels.
For a teacher model trained with standard (ground truth) labels $y$,
distillation fits a student model by minimizing a per-sample loss taken to
be a convex combination of the losses between the student's predicted label
$y_S$
and either the teacher's predicted label $y_T$ or the standard label $y$,
\begin{align*}
  \xi \cdot \ell(y_T, y_S) + (1-\xi)\cdot \ell(y, y_S)
\end{align*}
where $\ell$ is a loss function (e.g., squared error, negative log
likelihood) and $\xi \in [0,
  1]$.
The constant $\xi$ influences the degree to which the student attempts to
match the teacher's label, and in our analyses we take $\xi=1$, the
\emph{pure knowledge distillation} setting.

To formalize our model for knowledge distillation, we first define the
relevant spaces. Let the input-label space be $\mathcal{X}\times
\mathcal{Y}$, and let the student input space $\mathcal{V}$ be identified
with a subspace of $\mathcal{X}$.
Denote the teacher and student model parameter spaces by $\mathcal{T}$ and
$\mathcal{S}$, respectively.
The teacher model, which we assume is given, minimizes an objective
$L_T : \mathcal{T} \rightarrow \mathbb{R}$.

The teacher generates distilled labels via the mapping $h: \mathcal{T}
\times \mathcal{X} \rightarrow \mathcal{Y}'$, which represents the
prediction on input $x \in \mathcal{X}$.
The student model then minimizes an objective function $L_S : \mathcal{S}
\rightarrow \mathbb{R}$, which typically aggregates per-sample losses on the
domain $\mathcal{S} \times \mathcal{V} \times (\mathcal{Y} \cup
\mathcal{Y}')$ over the dataset.

\subsubsection{Related Work}

Along with studying distillation's influence on student
generalization\footnote{See Appendix~\ref{appendix:related-works-cont} for
additional related works.}, both theoretical and empirical works have also
analyzed its algorithmic effects. \citet{PL19} consider distillation applied
to linear models under gradient flow, the continuous time version of
gradient descent. They show that distillation biases the optimization
trajectory towards favorable, low generalization error solutions and that
data geometry influences the rate of convergence.

\citet{PLMRG25} identify a mechanism by which progressive distillation, where labels are produced by increasingly capable teacher models, accelerates the student's training. They prove sample complexity bounds for sparse parity and empirically test this phenomenon on language models for probabilistic context-free grammars and other natural language modeling tasks.

\citet{SPA23} analyze stochastic gradient descent under self-distillation and under distillation with unbiased compression operators. They show that distillation acts as a form of partial variance reduction, with the degree of reduction governed by the teacher's proximity to the optimum, and they lift this observation to upper bounds on the expected optimization error. In contrast, our analysis also considers students with strictly fewer parameters than the teacher, and we quantify the relative efficiency of knowledge distillation over supervised learning. Moreover, in the setting of rank-1 matrix approximation, we establish both upper and lower bounds, thereby demonstrating a concrete convergence gap.

\subsection{MLE in Generalized Linear Models}
We analyze stochastic gradient descent for generalized linear models \citep{NW72} fit with maximum likelihood estimation. In a GLM, conditional on the covariate $x \in \mathbb{R}^p$, the response $y \in \mathbb{R}$ follows a distribution in the exponential family. Specifically, the conditional density takes the form,
\begin{align*}
    p_\theta(y | x) = \exp\left(\frac{y(x^T\theta) - b(x^T\theta)}{a(\phi)} + c(y, \phi)\right).
\end{align*}
where $\theta \in \mathbb{R}^p$ is the true parameter, $\phi > 0$ is a constant, and $a$, $b$, and $c$ are known functions. The function $a(\phi)$ controls the dispersion, which we assume $a(\phi) = 1$. As written, the model uses the canonical link function, $g(\mathbb{E}[y | x]) = x^T\theta$, with inverse $h = g^{-1}$ mapping the linear predictor to the conditional mean of the response.

We fit the GLM by maximum likelihood estimation, so the per-sample loss function is the negative log-likelihood. With a slight overload of notation, we define the teacher $\ell_T$ and student $\ell_S$ per-sample losses as,
\begin{align*}
    \ell_T(\theta; x, y) &= -\log p_\theta(x | y), \\\ell_S(\gamma; v, y) &= -\log p_\gamma(v | y),
\end{align*}
where $\theta$, $x \in \mathbb{R}^p$ and $\gamma$, $v \in \mathbb{R}^d$ for $0 < d < p$. Data points $(x, y) \sim^{iid} \mathcal{P}$ come from a distribution over the input-label space, and $\{(x_i, y_i\}_{i=1}^n$ forms finite sample of size $n$. We define the projection onto the top-$d$ coordinates as $\Pi = [I_d \ 0]$, so the student observes $v = \Pi x$. The teacher's population and empirical objectives are
\begin{align*}
    L_T(\theta) &= \mathbb{E}[\ell_T(\theta; x, y)],\\ L_T^n(\theta) &= \frac{1}{n}\sum_{i=1}^n\ell_T(\theta; x_i, y_i),
\end{align*}
and the student's corresponding objectives are
\begin{align*}
    L_S(\gamma) &= \mathbb{E}[\ell_S(\gamma; v, y)], \\ L_S^n(\gamma) &= \frac{1}{n} \sum_{i=1}^n \ell_S(\gamma; v_i, y_i).
\end{align*}
Let $\theta^\ast$ (or $\hat{\theta}$) denote a teacher model minimizing its objective, population risk $L_T$ (or empirical risk $L_T^n$). Under knowledge distillation, the student observes the teacher's prediction $y' = h(x^T\theta^\ast)$ (or $y' = h(x^T\hat{\theta})$) as opposed to the standard label $y$. The distillation loss retains the form of $\ell_S$, but we replace the observed response with the teacher's prediction. In the population setting, this prediction satisfies $y' = \mathbb{E}[y | x]$, the conditional expectation of the response.

\subsubsection{Notation}
We use $\alpha$ to denote the student estimator fit with true labels $y$, and $\beta$ for the one fit with distillation labels $y'$. When referring to either estimator in a context where distinction is irrelevant, we use $\gamma$. Let the input-label pair be $z = (v, y)$ and $z' = (v, y')$. Use $L'_S(\gamma)$ and ${L'}_S^{n}(\gamma)$ to denote the student population and empirical objectives with distillation labels.

\subsection{Low Rank Approximation of a Matrix}
We then study the problem of finding a low-rank approximation of a positive semi-definite matrix $M \in S_+^n$. Specifically, we seek the rank-$r$ matrix (for $0 < r < n$) that minimizes the Frobenius norm of its difference from $M$. The solution corresponds to truncating the spectral decomposition of $M$ to its top-$r$ eigenvalues. \citet{BM03} introduce a nonconvex formulation that implicitly enforces both the rank and positive semi-definiteness constraints by optimizing over the factorized form $XX^T$ for ``tall and skinny'' matrices $X$. We define the teacher $L_T$ and student $L_S$ objective functions as,
\begin{align*}
    L_T(X) &= \frac{1}{4}||XX^T - M||_F^2,\\
    L_S(x) &= \frac{1}{4}||xx^T - M||_F^2
\end{align*}
where $X \in \mathbb{R}^{n \times r}$ and $x \in \mathbb{R}^n$. The teacher $X^\ast$ minimizes $L_T$, and generates targets via the mapping $h: \mathbb{R}^{n \times r} \rightarrow S_+^n$ defined by $h(X) = XX^T$. In the distillation setting, we replace the original matrix $M$ with $h(X^\ast)$ in the student's objective.

Note that the student objective $L_S$ has critical points at $\pm \sqrt{\lambda_i} v_i$, where $(\lambda_i, v_i)$ are eigenvalue-eigenvector pairs of either $M$ or $h(X^\ast)$, depending on the target. Gradient descent avoids spurious critical points, since the Hessian at such points has a direction of negative curvature. This follows from results by \citet{LSJR16,PP17}, which we review in our proof of Theorem~\ref{theorem:rank-1-convergence-to-mins}.

\subsubsection{Notation}
Eigenvalues of $M$ are ordered $\lambda_1 > \lambda_2 \ge \ldots \ge \lambda_n \ge 0$. The spectral gap $\Delta_i:= \lambda_1 - \lambda_i$ for $i=2, \ldots, n$. Let $M' = h(X^\ast)$ denote the distillation target. We use $w$ to denote the student estimator with target matrix $M$, and $u$ for the student with teacher-generated target $M'$. When referring to both, we use $x$. Let $L_S'(x)$ denote the objective function with target matrix $M'$.

\section{GENERALIZED LINEAR MODELS}\label{section:glm}
In this setting, we aim to recover the unique maximizer $\gamma^\ast$ of the likelihood function of a generalized linear model. We study how knowledge distillation influences stochastic gradient descent in this problem.

We defer all proofs from this section to Appendices~\ref{appendix:glm-sgd-online} and \ref{appendix:glm-sgd-finite}.

\subsection{Online Setting}\label{subsection:mle-online-sgd}
In this section, we focus on the following online stochastic gradient descent (SGD) update rule. At step $t$, given a sample $(x_t, \tilde{y}_t) \sim \mathcal{P}$ with $v_t = \Pi x_t$ and $\tilde{y}_t \in \{y_t, y_t'\}$, we update
\begin{align}\label{eq:sgd-update}
    \gamma_{t} = \gamma_{t-1} - \eta_t \nabla \ell_S(\gamma_{t-1}; v_t, \tilde{y}_t)
\end{align}
starting from $\gamma_0 \in \mathbb{R}^d$. We then analyze the dynamics of the averaged iterate,
\begin{align*}
    \bar{\gamma}_t = \frac{1}{t} \sum_{k=0}^{t-1} \gamma_k
\end{align*}
which achieves the optimal rate of $O(1/\sqrt{n})$ \citep{PJ92}. To establish consistency and asymptotic normality of $\bar{\gamma}_t$, we rely on the following assumptions. We state a simplified version of the assumption below, the full technical conditions are deferred to Appendix~\ref{appendix:glm-pop-assumptions}.

\begin{assumption}[Smoothness and convexity]\label{assumption:glm-pop-loss}
    The population losses $L_S$, $L_S'$ are strongly convex around their minimizers $\alpha^\ast$ and $\beta^\ast$, respectively. The per-sample losses $\ell_S(\alpha; z)$ and $\ell_S(\beta; z')$ are twice differentiable with locally Lipschitz gradients and Hessians, and their gradients have bounded variance.
\end{assumption}

\begin{assumption}[Step sizes]\label{assumption:pj-step-size}
    The sequence of step sizes satisfy $(\eta_t - \eta_{t+1})/\eta_t = o(\eta_t)$, $\eta_t > 0$ for all $t$, and
    \begin{align*}
        \sum_{t=1}^\infty \eta_t \cdot  t^{-1/2} < \infty.
    \end{align*}
\end{assumption}

\begin{theorem}[Population convergence]\label{theorem:glm-as-convergence-pop}
    Under Assumptions~\ref{assumption:glm-pop-loss} and \ref{assumption:pj-step-size}, let $\bar{\alpha}_t$ denote the averaged iterate from SGD with gradient estimates $\nabla \ell_S(\alpha_t; v_t, y_t)$, and let $\bar{\beta}_t$ denote the averaged iterate from SGD with $\nabla \ell_S(\beta_t; v_t, h(x_t^T \theta^\ast))$. Then both sequences converge almost surely to
    \begin{align*}
        \gamma^\ast := \argmin_{\gamma \in \mathbb{R}^d} \ L_S(\gamma)
    \end{align*}
\end{theorem}

Having established that supervised learning and knowledge distillation converge to the same limiting parameter, we now show that the SGD iterates obtained from knowledge distillation, $\bar{\beta}_t$, are asymptotically more efficient than the ones obtained from supervised learning, $\bar{\alpha}_t$.

\begin{theorem}\label{theorem:glm-asymp-normality-pop}
    Suppose Assumptions \ref{assumption:glm-pop-loss} and \ref{assumption:pj-step-size} hold. Let $\gamma^\ast$ be the minimizer of $L_S$, and define the Hessian at $\gamma^\ast$
    \begin{align*}
        H = \mathbb{E}[h'(v^T\gamma^\ast)vv^T]
    \end{align*}
    where $h$ is the inverse link function and $v = \Pi x$. Define the second moments of the gradient noise at $\gamma^\ast$,
    \begin{align*}
        C(\gamma^\ast) &= \mathbb{E}[(h(v^T\gamma^\ast) - y)^2 vv^T], \\
        C'(\gamma^\ast) &= \mathbb{E}[(h(v^T\gamma^\ast) - h(x^T \theta^\ast))^2 vv^T]
    \end{align*}
    Let $\bar{\alpha}_t$ be the averaged iterate under supervised learning, obtained from SGD with gradient estimates $\nabla \ell_S(\alpha_t; v_t, y_t)$. Let $\bar{\beta}_t$ be the averaged iterate under knowledge distillation, obtained from SGD with $\nabla \ell_S(\beta_t; v_t, h(x_t^T \theta^\ast))$. Then as $t \rightarrow \infty$,
    \begin{align*}
        \sqrt{t}(\bar{\alpha}_t - \gamma^\ast) \overset{d}{\longrightarrow} \mathcal{N}(0, H^{-1} C(\gamma^\ast) H^{-1}).
    \end{align*}
    and
    \begin{align*}
        \sqrt{t}(\bar{\beta}_t - \gamma^\ast) \overset{d}{\longrightarrow} \mathcal{N}(0, H^{-1}C'(\gamma^\ast)H^{-1}).
    \end{align*}
    Moreover, the asymptotic covariance under knowledge distillation is no larger than under supervised learning,
    \begin{align*}
        C(\gamma^\ast) - C'(\gamma^\ast) \succeq \mathbb{E}[\text{Var}(y | x) vv^T] \succeq 0,
    \end{align*}
    which holds with equality only if the label variance is zero.
\end{theorem}

By Theorem 2 of \citet{PJ92}, both $\bar{\alpha}_t$ and $\bar{\beta}_t$ converge almost surely $\gamma^\ast$ and are asymptotically normal. Assumptions~\ref{assumption:glm-pop-loss} and~\ref{assumption:pj-step-size} provide the technical conditions under which this classical stochastic approximation result applies.

The key intuition is that projecting onto the teacher's manifold before projecting onto the student's manifold does not change the final point. For example, in linear regression (identity link) with Gaussian noise, the maximum likelihood estimator corresponds to the orthogonal projection of the response onto the span of the covariates. Projecting first onto the span of the full $p$-dimensional covariates and then onto the span of the truncated $d$-dimensional covariates produces the same results as projecting directly onto the truncated covariates.

Knowledge distillation improves efficiency, $\bar{\beta}_t$ has a smaller asymptotic covariance than $\bar{\alpha}_t$, by removing the label variance component from the stochastic gradient estimate. The efficiency gain is most pronounced when the response has high variance, either due to label noise or the intrinsic variability of the distribution, while the benefit diminishes when the labels are nearly deterministic and individual samples already provide reliable information.

\subsection{Finite Dataset Setting}\label{subsection:mle-finite-dataset-sgd}
The online SGD setting in the previous section samples a new observation at each step $t$, analogous to training with a single pass over the data. Here, we instead assume access to a finite dataset $D = \{(x_i, y_i)\}_{i=1}^n \sim \mathcal{P}^n$, where pairs are drawn i.i.d. from $\mathcal{P}$. Once realized, we treat $D$ as fixed and compute each SGD update (Eq.~\ref{eq:sgd-update}) using a sample $(x_t, y_t) \sim U(D)$ chosen uniformly from $D$.

Under standard regularity conditions, the averaged iterates $\bar{\gamma}_t$ are consistent and asymptotically normal. Their behavior parallels the conclusions of Theorems~\ref{theorem:glm-as-convergence-pop} and \ref{theorem:glm-asymp-normality-pop}, except that the empirical Hessian and gradient noise second moments replace their population counterparts.

We present simplified versions of the assumptions below. The full technical conditions are deferred to Appendix~\ref{appendix:glm-emp-assumptions}.

\begin{assumption}[Smoothness and convexity]\label{assumption:glm-emp-loss}
    The empirical losses $L_S^n$, ${L'}_S^n$ are strongly convex around their minimizers $\hat{\alpha}$ and $\hat{\beta}$, respectively. The per-sample losses $\ell_S(\alpha; z)$ and $\ell_S(\beta; z')$ are twice differentiable with locally Lipschitz gradients and Hessians, and their gradients have bounded growth.
\end{assumption}

\begin{assumption}[Concentration]\label{assumption:emp-concentration}
    The covariates and labels satisfy standard regularity conditions, ensuring concentration of empirical quantities around their population counterparts.
    
    In particular, the conditional variance of $y$ given $x$ is uniformly bounded away from zero and per-sample gradient covariance matrices are uniformly bounded around $\gamma^\ast$.
\end{assumption}

\begin{theorem}[Convergence under finite dataset]\label{theorem:glm-as-convergence-emp}
    Under Assumptions~\ref{assumption:glm-emp-loss} and \ref{assumption:pj-step-size}, suppose we have a fixed dataset $D = \{(x_i,y_i)\}_{i=1}^n$, and at each iteration $t$ the sample $(x_t,y_t)$ is drawn uniformly at random from $D$ (with replacement).
    
    Let $\bar{\alpha}_t$ denote the averaged iterate from SGD with gradient estimates $\nabla \ell_S(\alpha_t; v_t, y_t)$, and let $\bar{\beta}_t$ denote the averaged iterate from SGD with $\nabla \ell_S(\beta_t; v_t, h(x_t^T\hat{\theta}))$. Then both sequences converge almost surely to
    \begin{align*}
        \hat{\gamma} := \argmin_{\gamma \in \mathbb{R}^d} \ L_S^n(\gamma).
    \end{align*}
\end{theorem}

As the number of SGD steps tends to infinity, both supervised learning $\bar{\alpha}_t$ and knowledge distillation $\bar{\beta}_t$ tend to the same minimizer of the empirical loss.

\begin{theorem}\label{theorem:glm-asym-normality-emp}
    Suppose Assumptions~\ref{assumption:glm-emp-loss}, \ref{assumption:emp-concentration}, and \ref{assumption:pj-step-size} hold. Let $\hat{\gamma}$ be the minimizer of the empirical loss $L_S^n$, and define the empirical Hessian at $\hat{\gamma}$,
    \begin{align*}
        H_n = \frac{1}{n} \sum_{i=1}^n h'(v_i^T\hat{\gamma})v_iv_i^T
    \end{align*}
    where $h$ is the inverse link function and $v_i= \Pi x_i$. Define the empirical second moments of gradient noise at $\hat{\gamma}$,
    \begin{align*}
        C_n(\hat{\gamma}) &= \frac{1}{n}\sum_{i=1}^n (h(v_i^T \hat{\gamma}) - y_i)^2v_iv_i^T, \\
        C_n'(\hat{\gamma}) &= \frac{1}{n} \sum_{i=1}^n (h(v_i^T \hat{\gamma}) - h(x_i^T \hat{\theta}))^2 v_iv_i^T
    \end{align*}
    where $\hat{\theta}$ is the teacher estimator. 
    
    Let $\bar{\alpha}_t$ denote the averaged iterate under supervised learning, obtained from SGD with gradient estimates $\nabla \ell_S(\alpha_t; v_t, y_t)$. Let $\bar{\beta}_t$ denote the averaged iterate under knowledge distillation, obtained from SGD with $\nabla \ell_S(\beta_t; v_t, h(x_t^T \hat{\theta}))$. Then as $t\rightarrow \infty$,
    \begin{align*}
        \sqrt{t}(\bar{\alpha}_t - \hat{\gamma}) \overset{d}{\longrightarrow} \mathcal{N}(0, H_n^{-1} C_n(\hat{\gamma}) H_n^{-1})
    \end{align*}
    and
    \begin{align*}
        \sqrt{t}(\bar{\beta}_t - \hat{\gamma}) \overset{d}{\longrightarrow} \mathcal{N}(0, H_n^{-1} C_n'(\hat{\gamma})H_n^{-1}).
    \end{align*}
    Moreover, for sufficiently large $n$,
    \begin{align*}
        C_n(\hat{\gamma}) - C_n'(\hat{\gamma}) \succeq 0
    \end{align*}
    with high probability. 
\end{theorem}

Analogous to the online SGD (population) setting, almost sure convergence of $\bar{\alpha}_t$ and $\bar{\beta}_t$ to $\hat{\gamma}$, together with their asymptotic normality, follows from Theorem 2 of \citet{PJ92}.

In the population case, the relative efficiency gap between knowledge distillation and supervised learning increases with the label variance. Theorem~\ref{theorem:glm-asym-normality-emp} establishes this relationship in the finite-dataset setting as well. Since the empirical second moments of the gradient noise are evaluated at the empirical minimizer $\hat{\gamma}$ rather than the population minimizer $\gamma^\ast$, the argument alternates between applying local Lipschitz continuity of the empirical second moments and using concentration bounds to control their deviation from the population counterparts at $\gamma^\ast$.

Thus, the efficiency of knowledge distillation from the population setting carries over to finite datasets. Under high label variance, distillation provides a tangible improvement, while in low-variance regimes its effect is negligible.

\section{LOW RANK APPROXIMATION}\label{section:low-rank}
In this section, we study the problem of finding the best rank-1 approximation of a positive semi-definite matrix $M \in S_+^n$. We formulate this as a nonconvex optimization problem \citep{BM03} and investigate how knowledge distillation influences the convergence behavior of gradient descent.

We defer all proofs from this section to Appendix~\ref{appendix:rank-1}.

\subsection{Convergence Analysis}\label{subsection:low-rank-convergence}
We analyze the convergence of gradient descent with constant step size $\eta$. At each step, we perform the update
\begin{align*}
    x_{t+1} = x_t - \eta \nabla \widetilde{L}_S(x_t).
\end{align*}
where $\widetilde{L}_S \in \{L_S, L_S'\}$. While the gradients $\nabla L_S$ and $\nabla L_S'$ differ, the update takes the same general form.

\begin{theorem}\label{theorem:rank-1-convergence-to-mins}
    Let the step size be
    \begin{align*}
        0 < \eta < \frac{1}{L \cdot (\lambda_1 + 1)}
    \end{align*}
    where $\lambda_1$ is the largest eigenvalue of the target matrix $M$ and $L \ge 4$ is a fixed constant. Sample $n_0 \sim \mathcal{N}(0, k\lambda_1/n \cdot I_n)$ with some $k \in (0, 1)$, and condition on $||n_0||_2 \le \sqrt{\lambda_1}$. Set the initial points of the gradient descent trajectories to be equal, $w_0 = u_0 = n_0$.
    
    Let $w_t$ denote the iterate from gradient descent on standard labels, $L_S$, and let $u_t$ denote the iterate from distillation labels, $L_S'$. Then both sequences converge almost surely to
    \begin{align*}
        x^\ast \in \argmin_{x \in \mathbb{R}^n} L_S(x).
    \end{align*}
\end{theorem}

When $r > 1$, the set of critical points of $L_S$ and $L_S'$ is strictly larger than the set of all global minimizers. In fact, their gradients are zero at any point of the form $\pm \sqrt{\lambda_i} v_i$ for $i \in [n]$ (or $i \in [r]$). To establish that the sequences $\{w_t\}$ and $\{u_t\}$ converge to a minimizer rather than a strict saddle, we show that an open convex set is forward-invariant under the gradient updates and then apply Theorem 3 of \citet{PP17}.

Consequently, the limiting matrices satisfy $w^\ast(w^\ast)^T = u^\ast(u^\ast)^T$, even though the limit points $w^\ast$ and $u^\ast$ themselves may differ. We now analyze the gradient descent dynamics over a finite number of steps, focusing on the number of iterations required to achieve a prescribed error tolerance.

\begin{definition}
    For a tolerance $\varepsilon > 0$, define the first step when the gradient descent sequence $\{x_t\}$ enters the $(\lambda_1\cdot\varepsilon)$-ball around the optimum $x^\ast$ as
    \begin{align*}
        T(x) := \min\{t: ||x_t - x^\ast||_2^2 < \lambda_1 \cdot \varepsilon\}
    \end{align*}
\end{definition}
\begin{definition}\label{def:initial-imbalance}
    For an iterate $x_t = (x_{1,t}, x_{2:n,t})$ expressed in the eigenbasis of $M$, define
    \begin{align*}
        \textsc{Im}(x_t) := \frac{||x_{2:n,t}||_2^2}{x_{1,t}^2}.
    \end{align*}
\end{definition}

Equivalently, $\sqrt{\textsc{Im}(x_t)}$ is the tangent of the angle between $x_t$ and the top eigenvector, so $\textsc{Im}(x_t)$ quantifies how misaligned the iterate is with the leading direction.
    
\begin{lemma}[Supervised learning\footnote{Appendix~\ref{appendix:rank1-lower-bound} contains a detailed statement of the lemma together with its proof.}]\label{lemma:rank1-lower-bound}
    Given $\varepsilon > 0$, under the previous step size and random initialization conditions, the number of steps required by supervised learning satisfies a lower bound of the form
    \begin{align*}
        T(w) \ge \frac{1}{2\log(1/(1-\eta\Delta_n))}\log\left(\frac{\textsc{Im}(x_0)}{\varepsilon}\right).
    \end{align*}
\end{lemma}
\begin{lemma}[Knowledge distillation\footnote{Appendix~\ref{appendix:rank1-upper-bound} contains a detailed statement of the lemma together with its proof.}]\label{lemma:rank1-upper-bound}
    Given $\varepsilon > 0$, under the same conditions, the number of iterations required by knowledge distillation is bounded above by 
    \begin{align*}
        T(u) \le \max\{t_a, t_b, t_\xi\} + t_\delta,
    \end{align*}
    where $t_a,t_b,t_\xi,t_\delta$ depend logarithmically on the initialization $\textsc{Im}(x_0)$, tolerance, and spectrum. 
\end{lemma}

The lower and upper bounds on $T(w)$ and $T(u)$ allow us to establish conditions under which gradient descent with the knowledge distillation objective converges faster than supervised learning.

\begin{theorem}\label{theorem:conditions-for-kd-gd-faster}
    Given $\varepsilon > 0$ and teacher matrix rank $r$, let the step size be
    \begin{align*}
        0 < \eta < \frac{1}{L \cdot (\lambda_1 + 1)}
    \end{align*}
    where $\lambda_1$ is the largest eigenvalue of the target matrix $M$ and $L \ge 4$ is a fixed constant. Sample $n_0 \sim \mathcal{N}(0, k\lambda_1/n \cdot I_n)$ with some $k \in (0, 1)$. Set the initial points of the gradient descent trajectories to be equal, $w_0 = u_0 = n_0$. 

    Assume furthermore that $\Delta_n < \ell\lambda_1$ and $(1+\eta\lambda_1/2)(1-\eta\Delta_n) > 1$ with $\ell = L/(L+1)$. Then there exists $c > 0$ (depending on the spectrum of $M$) such that for all $n = \Omega(\varepsilon^{-c})$, we have $T(w) - T(u) \ge 0$ with constant probability over the initialization.
\end{theorem}

The requirement that the dimension $n$ grows as $\Omega(\varepsilon^{-c})$ may seem undesirable, but it also not unexpected. The top spectral gap $\Delta_2$ controls the asymptotic rate of gradient descent, leading to a $O(1/\Delta_2)$ dependence. Knowledge distillation cannot improve this rate unless the teacher itself has rank $r=1$, in which case the dependence reduces to $O(1/\lambda_1)$. Lemma~\ref{lemma:rank1-upper-bound} shows that the main advantage of knowledge distillation lies in accelerating the decay of coordinates $r+1$ through $n$. As the target tolerance $\varepsilon$ decreases, maintaining this advantage requires increasing mass in these trailing coordinates.

\begin{figure}[t]
    \centering
    \includegraphics[width=0.85\linewidth]{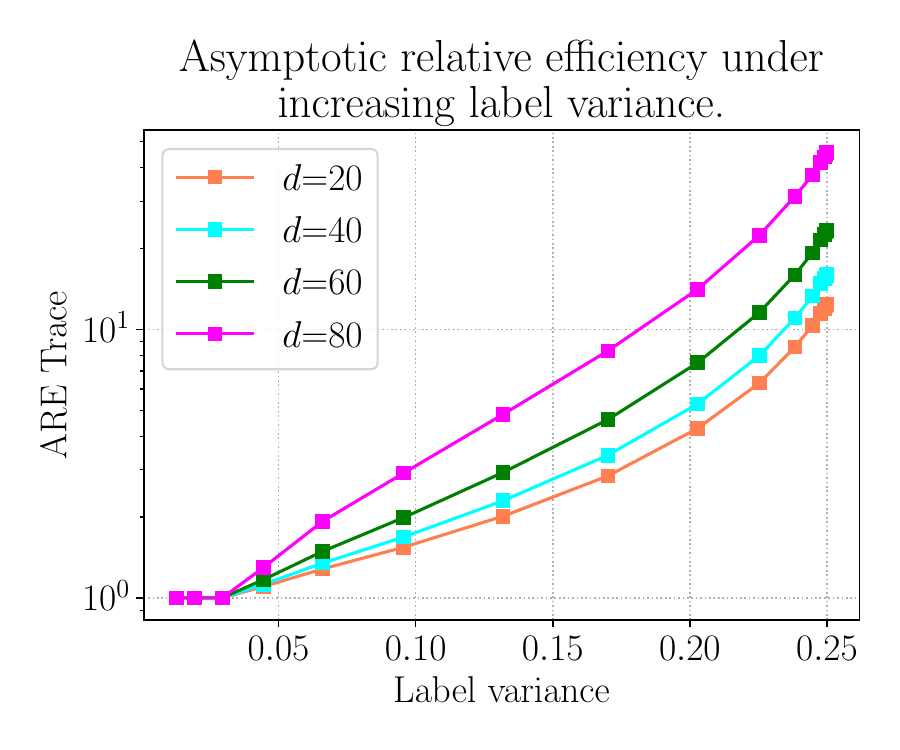}
    \caption{Asymptotic relative efficiency, given by the ratio of asymptotic covariance traces (supervised learning over knowledge distillation), as a function of label variance. The teacher dimension is fixed at $p=1000$, and the student dimension is $d$. Each point represents an average over 100 sampled datasets and true parameters.}
    \label{fig:theoretical-asymp-covariance}
\end{figure}

\section{EXPERIMENTS}\label{section:experiments}
We test our theory on both synthetic and real data. The code for our experiments can be found at \href{https://github.com/0aax/two-models-kd}{https://github.com/0aax/two-models-kd}.

\subsection{Synthetic Experiments}\label{subsection:synthetic-experiments}

\begin{figure}[t]
    \centering
    \includegraphics[width=0.85\linewidth]{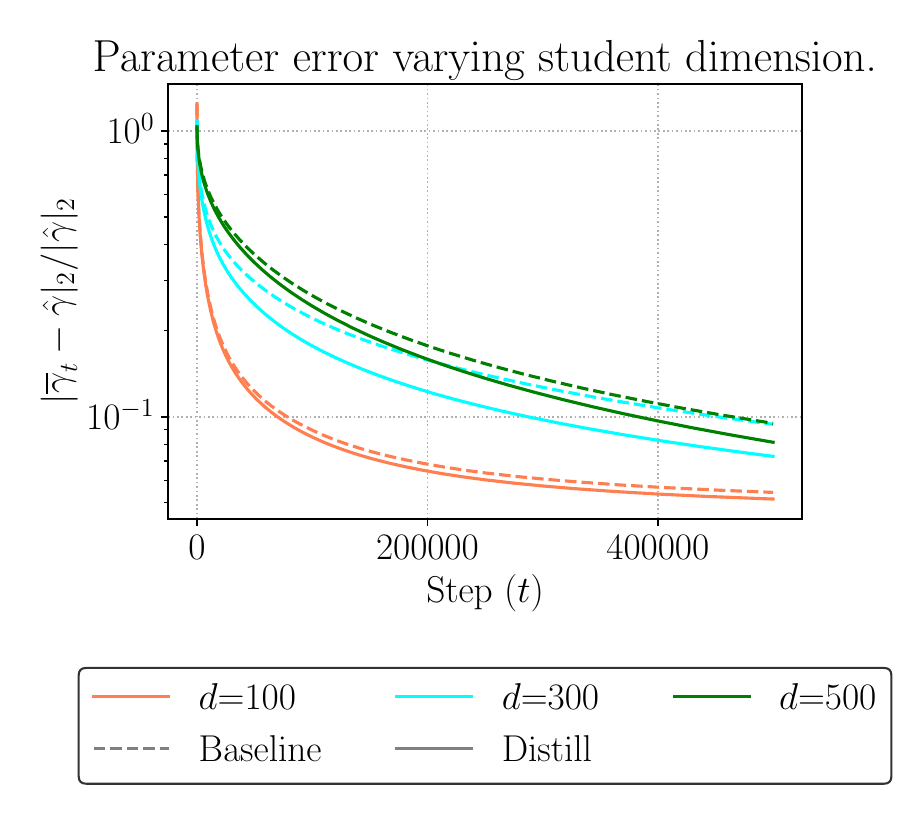}
    \caption{(Multinomial logistic regression, Fashion-MNIST) Normalized parameter error of SGD iterates under knowledge distillation and standard supervision for student dimensions $d \in\{100, 300, 500\}$. The initial step size $\eta_0$, optimal for standard supervision, is used for both methods. Results are averaged over 10 trials.}
    \label{fig:fmnist}
\end{figure}

\paragraph{Generalized linear models.}
For generalized linear models, our theory predicts that distillation improves efficiency by removing label variance from the gradient estimates. To test this prediction, we construct a logistic regression dataset with $n=1000$ samples and dimension $p=100$ covariates, drawn i.i.d. from $\mathcal{N}(0, 1/p\cdot I_p)$. Labels follow the Bernoulli distribution with mean $\sigma(x_i^T \theta^\ast)$, where the scale of the true parameter $\theta^\ast = \kappa \cdot n_0$ (with $n_0 \sim \mathcal{N}(0, I_p)$, $\kappa \in [10^{-1}, 10^{2.5}]$) controls the label variance. We apply L-BFGS \citep{LN89} to obtain the empirical minimizers $\hat{\alpha}$ and $\hat{\beta}$, and use them to compute the asymptotic covariance matrices of the supervised and distilled estimators. Figure~\ref{fig:theoretical-asymp-covariance} shows that distillation achieves higher relative efficiency as label variance and student dimension $d \in \{20, 40, 60, 80\}$ increase.

\begin{figure*}[t]
    \centering
    \includegraphics[width=0.95\linewidth]{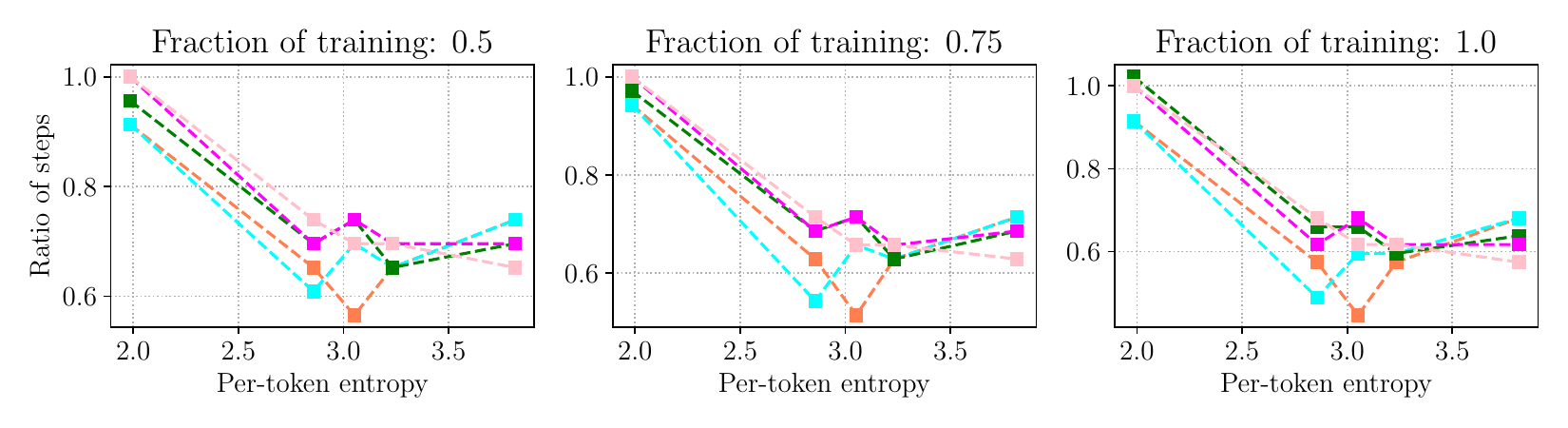}
    \caption{(Language model pretraining) Ratio of training steps required by knowledge distillation versus standard supervision across subsets of increasing entropy, evaluated at 0.5, 0.75, and 1.0 fractions of training steps. At each point, we record the maximum of the two validation losses, and measure how many steps each student needs to reach this loss. This is repeated over five trials.}
    \label{fig:c4-plot}
\end{figure*}

We then ask whether the asymptotic efficiency translates into fewer optimization steps. We construct the dataset in the manner detailed above, with $p=10$ and $\kappa=1.0$, and fit the teacher $\hat{\theta}$ using L-BFGS. Both supervised and distilled students start at the same point $\alpha_0 = \beta_0 = 0$ and we run SGD with step sizes $\eta_t = \eta_0/(1+10^{-3} t^{0.6})$ where $\eta_0 \in \{10^{-3}, 10^{-2.5}, \ldots, 10^{-0.5}, 1.0\}$, consistent with Assumption~\ref{assumption:pj-step-size}. Figure~\ref{fig:glm-synthetic-both} (Appendix~\ref{appendix:glm-synthetic}) shows that, even when both methods use the same $\eta_0$ (optimal for standard supervision), knowledge distillation converges faster. Furthermore, this gap becomes more significant when distillation uses its own optimal $\eta_0$.

\paragraph{Rank-1 matrix approximation.}

Our analysis in Section~\ref{section:low-rank} shows that distillation accelerates convergence when the teacher truncates a ``heavy tail'' of eigenvalues (i.e. when the removed eigenvalues are, in sum, relatively large compared to the top-$r$ retained eigenvalues). To test this, we construct a diagonal target matrix of size $n=500$ with a controlled spectrum, $M = \mathrm{diag}(\lambda_1, s_r, s_n)$, for arithmetic sequences $s_r$ and $s_n$. We obtain the teacher matrix by truncating to the top-$r$ eigenvalues, equivalent to zeroing out entries $r+1$ through $n$ on the diagonal. 

Gradient descent starts at $x_0 \sim \mathcal{N}(0, 1/n \cdot I_n)$ and we measure progress via the distance to the global optimum $\pm \sqrt{\lambda_1} e_1$. We sweep step sizes $\eta \in \{10^{-2}, 10^{-1.5}, 10^{-1}, 10^{-0.5}, 1.0\}$ and select one which minimizes parameter error for supervised learning. In all settings, this is $\eta = 10^{-0.5}$. Figure~\ref{fig:rank-1-both} (Appendix~\ref{appendix:rank1-synthetic}) shows that distillation accelerates convergence when the teacher rank approaches the student's rank and when it truncates relatively large eigenvalues.

\subsection{Regression on Real Data}
We next test our predictions on real-world data. On Fashion-MNIST~\citep{XRV17}, we fit multinomial logistic regression models by flattening each $28\times 28$ image decomposing it into its principal components. The teacher observes to the full set of components, while the student observes only a subset of the components ($d \in \{100, 300, 500\}$). Both supervised and distilled students start at the same initialization $\alpha_{0}^c = \beta_0^c = n_0^c$, where $n_0^c \sim \mathcal{N}(0, 1/d \cdot I_d)$ and $c = 1, \ldots, 10$. The batch size is $b=100$ and step size $\eta_t = \eta_0/(1+10^{-3} t^{0.6})$, consistent with Assumption~\ref{assumption:pj-step-size}. We sweep step sizes $\eta_0 \in \{10^{-1}, 10^{-0.75}, 10^{-0.5}, 10^{-0.25}, 1.0\}$ and select the one which has the lowest parameter error for standard supervision. This ends up being: $\eta_0=10^{-0.75}$ for $d=100$, $\eta_0=10^{-0.5}$ for $d=300$, and $\eta_0=10^{-0.5}$ for $d=500$. Figure~\ref{fig:fmnist} shows the normalized distance to the optimum over training steps. Distillation consistently converges in fewer steps across all student dimensions. We find the optimal step size $\eta_0$ for distillation to be larger in some cases than the optimal for standard supervision.\footnote{Figure~\ref{fig:fmnist-best-lr} (Appendix~\ref{appendix:fmnist-best-lr}) shows error under optimal $\eta_0$ for both methods.}

\subsection{Next Token Prediction}
Finally, we move beyond simple regression and matrix approximation to study language model pretraining. While our GLM analysis suggests that distillation benefits setting with high label variance, since we cannot directly observe label variance in next-token prediction, we use per-token entropy of a model's predicted distribution as a proxy. We train a GPT-2 medium model on a subset of the C4 dataset~\citep{RSRL20} and compute per-token entropy to partition the remaining data into subsets of increasing entropy.

For each subset, we train a GPT-2 small student model with different learning rates and select the model with the lowest validation loss as the supervised learning baseline. We then train the distillation student model with the same training setup as the supervised learning baseline. The only difference in the two setups being the training objective, the baseline minimizes cross-entropy with ground-truth labels, while the distilled student minimizes the KL-divergence $D_{\mathrm{KL}}(P_T || P_S)$ between teacher $P_T$ and student $P_S$ distributions, where the teacher GPT-2 medium model trains on the same subset.\footnote{Appendix~\ref{appendix:pretraining-details} contains additional experimental details.}

Figure~\ref{fig:c4-plot} shows the ratio of training steps required by the student trained with knowledge distillation over the supervised learning baseline to reach a particular validation loss on each of the subsets that are increasing in entropy. Across all subsets, distillation reaches the target loss in fewer steps, with the improvement largest in high-entropy subsets.

\section{CONCLUSION}
In this work, we analyze knowledge distillation in two settings, generalized linear models fit with maximum likelihood estimation and rank-1 matrix approximation. For generalized linear models, we show that distillation and standard supervision converge to the same solution, but that distillation produces asymptotically more efficient SGD iterates in both online and finite-dataset settings. In rank-1 matrix approximation, we prove that both approaches converge to the same set of global optima and identified conditions under which distillation converges strictly faster. Our experiments on synthetic data, regression on real data, and language model pretraining corroborate our theory and suggests its practical relevance.

The limitations of our current analysis provide several avenues for future work. We focused on two setting with relatively benign geometry, extending this analysis to more complex optimization landscapes, where gradient descent may converge to suboptimal points, would more closely resemble practice. Another direction continues analyzing the GLM setting, but establishes non-asymptotic results. Lastly, extending empirical evaluations across a broader range of data domains may provide additional insights into how the underlying data distribution influences distillation.

\section{ACKNOWLEDGMENTS}
This work was partially supported by the Office for Naval Research under grant N00014-22-1-2669, NSF AI Institute for Foundations of Machine Learning (IFML), and Open Philanthropy. The authors would also like to thank Suhas Kotha and Gregory Xie for helpful feedback on the paper.

\bibliographystyle{plainnat}
\bibliography{references}

\section*{Checklist}

\begin{enumerate}

  \item For all models and algorithms presented, check if you include:
  \begin{enumerate}
    \item A clear description of the mathematical setting, assumptions, algorithm, and/or model. Yes.
    \item An analysis of the properties and complexity (time, space, sample size) of any algorithm. Not Applicable.
    \item (Optional) Anonymized source code, with specification of all dependencies, including external libraries. Yes.
  \end{enumerate}

  \item For any theoretical claim, check if you include:
  \begin{enumerate}
    \item Statements of the full set of assumptions of all theoretical results. Yes.
    \item Complete proofs of all theoretical results. Yes.
    \item Clear explanations of any assumptions. Yes. 
  \end{enumerate}

  \item For all figures and tables that present empirical results, check if you include:
  \begin{enumerate}
    \item The code, data, and instructions needed to reproduce the main experimental results (either in the supplemental material or as a URL). Yes.
    \item All the training details (e.g., data splits, hyperparameters, how they were chosen). Yes.
    \item A clear definition of the specific measure or statistics and error bars (e.g., with respect to the random seed after running experiments multiple times). Yes.
    \item A description of the computing infrastructure used. (e.g., type of GPUs, internal cluster, or cloud provider). Yes.
  \end{enumerate}

  \item If you are using existing assets (e.g., code, data, models) or curating/releasing new assets, check if you include:
  \begin{enumerate}
    \item Citations of the creator If your work uses existing assets. Yes.
    \item The license information of the assets, if applicable. Not Applicable.
    \item New assets either in the supplemental material or as a URL, if applicable. Not Applicable.
    \item Information about consent from data providers/curators. Not Applicable.
    \item Discussion of sensible content if applicable, e.g., personally identifiable information or offensive content. Not Applicable.
  \end{enumerate}

  \item If you used crowdsourcing or conducted research with human subjects, check if you include:
  \begin{enumerate}
    \item The full text of instructions given to participants and screenshots. Not Applicable.
    \item Descriptions of potential participant risks, with links to Institutional Review Board (IRB) approvals if applicable. Not Applicable.
    \item The estimated hourly wage paid to participants and the total amount spent on participant compensation. Not Applicable.
  \end{enumerate}

\end{enumerate}

\appendix
\thispagestyle{empty}

\onecolumn
\aistatstitle{Appendix}

\section{RELATED WORK CONTINUED}\label{appendix:related-works-cont}
Knowledge distillation has shown repeated success in training small performant models using larger, more capable models. Prior theoretical analyses that have sought to explain why, have done so primarily from the perspective that distillation yields student models with better generalization. \citet{MRRKK20} connects teacher predictions in classification to modeling the Bayes class probabilities. They show that empirical risk with distillation labels (Bayes-distilled risk) has lower variance relative to standard labels, and the student minimizing the Bayes-distilled risk inherits the variance reduction, leading to better generalization. \citet{DKSM21} views distillation as a plug-in method for semiparameteric inference, where the teacher predictions serve as a plug-in estimate of the Bayes class probabilities. They establish excess risk guarantees for the student that depend on teacher quality, and introduce loss correction to address teacher bias and cross-fitting to mitigate overfitting from data reuse.

\citet{MFB20} analyze self-distillation, where teacher and student are the same, in a Hilbert space with $\ell_2$ regularization. They show that repeated distillation shrinks basis coefficients more aggressively, thus acting as a strong regularizer that can reduce overfitting and improve generalization. \citet{DS23} provide another explanation for self-distillation. They consider a student that minimizes a dual objective on both original and teacher-produced labels, and characterize settings where distillation produces students with higher accuracy in ridge regression and binary logistic regression. \citet{PDO24} generalizes this to multistep self-distillation.

\citet{IGTMO25} analyze knowledge distillation in high-dimensional ridgeless linear regression with Gaussian covariates. They show that in the underparameterized regime ($n > p$) the optimal teacher coincides with the ground truth parameter, so distillation cannot outperform direct supervision, while in the overparameterized regime ($n < p$) an optimally chosen teacher reduces the risk of the student.

While these works study the student model defined as the minimizer of the objective under knowledge distillation, our focus is instead on the successive iterates generated by descent algorithms that approximate this minimizer.

\section{Detailed assumptions and deferred proofs for Section~\ref{subsection:mle-online-sgd}}\label{appendix:glm-sgd-online}

\subsection{Assumptions}\label{appendix:glm-pop-assumptions}
In this section, we present the formal assumptions used in Theorems~\ref{theorem:glm-as-convergence-pop} and \ref{theorem:glm-asymp-normality-pop}. These assumptions apply to both the standard supervision objectives, $\ell_S(\alpha; z)$ and $L_S(\alpha)$, and the knowledge distillation objectives, $\ell_S(\beta; z')$ and $L_S'(\beta)$. To streamline the presentation, we introduce the notation $\ell_S(\gamma; \tilde{z})$ and $\widetilde{L}_S(\gamma)$, with $\tilde{z} \in \{z, z'\}$, to state the assumptions that apply to both settings.

\begin{assumption}\label{assumption:glm-pop-loss-full}
    The per-sample loss $\ell_S(\gamma; \tilde{z})$ and population loss $\widetilde{L}_S(\gamma) = \mathbb{E}[\ell_S(\gamma; \tilde{z})]$ where $\tilde{z} = (v, \tilde{y})$, $v=\Pi x$, and $\tilde{y}\in\{y, y'\}$ satisfies the following:
    \begin{enumerate}
        \item[(i)] At the unique optimal point $\gamma^\ast := \argmin_\gamma \widetilde{L}_S(\gamma)$, for a constant $m > 0$, the Hessian $\nabla^2 \widetilde{L}_S(\gamma^\ast) \succeq m \cdot I_d$ is positive definite.
        \item[(ii)] The map $\gamma \mapsto \ell_S(\gamma; \tilde{z})$ is twice continuously differentiable for every $\tilde{z}$. There exists a random variable $M(\tilde{z})$ with $\mathbb{E}[M(\tilde{z})] < \infty$ and a constant $r > 0$ such that for all $\gamma \in B(\gamma^\ast, r)$, 
        \begin{align*}
            ||\nabla \ell_S(\gamma; \tilde{z}) - \nabla \ell_S(\gamma^\ast; \tilde{z})||_2 &\le M(\tilde{z}) \cdot ||\gamma - \gamma^\ast||_2 \\
            ||\nabla^2 \ell_S(\gamma; \tilde{z}) - \nabla^2 \ell_S(\gamma^\ast; \tilde{z})||_{op} &\le M(\tilde{z}) \cdot ||\gamma- \gamma^\ast||_2.
        \end{align*}
        \item[(iii)] There exists a constant $C > 0$ such that the per-sample gradient $\nabla \ell_S(\gamma; \tilde{z})$, is bounded $\mathbb{E}[||\nabla \ell_S(\gamma; \tilde{z})||_2^2] < C \cdot (1 + ||\gamma||_2^2)$.
    \end{enumerate}
\end{assumption}

We now specialize Theorem 2 of \citet{PJ92} to our setting of GLMs with the population objective and state it as a corollary.

\begin{corollary}[\citet{PJ92}, GLMs with objective $\widetilde{L}_S$] \label{corollary:pj-for-pop-glms}
    Consider the stochastic gradient descent update with sample $(v_t, \tilde{y}_t)$,
    \begin{align*}
        \gamma_{t} = \gamma_{t-1} - \eta_t \nabla \ell_S(\gamma_{t-1}; v_t, \tilde{y}_t).
    \end{align*}
    Assume \ref{assumption:glm-pop-loss-full} and \ref{assumption:pj-step-size} are true, then $\bar{\gamma}_t \rightarrow \gamma^\ast$ almost surely, and
    \begin{align*}
        \sqrt{t} (\bar{\gamma}_t - \gamma^\ast) \overset{d}{\longrightarrow} \mathcal{N}(0, \widetilde{H}^{-1} \widetilde{C}(\gamma^\ast) \widetilde{H}^{-1})
    \end{align*}
    with second moment of the gradient noise at the optimum $\widetilde{C}(\gamma^\ast) = \mathbb{E}[\nabla \ell_S(\gamma^\ast; v, \tilde{y})(\nabla \ell_S(\gamma^\ast; v, \tilde{y}))^T]$ and Hessian $\widetilde{H} = \nabla^2 \widetilde{L}_S(\gamma^\ast)$.
\end{corollary}

\begin{proof}
    Begin by considering the recursion
    \begin{align*}
        \gamma_t = \gamma_{t-1} - \eta_t g_t
    \end{align*}
    where $\xi_t = \nabla \ell_S(\gamma_{t-1}; v_t, \tilde{y}_t) - \nabla \tilde{L}_S(\gamma_{t-1})$ and $g_t = \nabla \tilde{L}_S(\gamma_{t-1}) + \xi_t$. Note that this is the stochastic optimization scheme analyzed in Theorem 2 of \citet{PJ92}. Now we show that the original assumptions are satisfied. For the assumptions needed to prove almost sure convergence of SGD using the theorem by \citet{RS71}, note that the gradient noise $\{\xi_t\}_{t \ge 1}$ is a martingale-difference sequence with respect to the natural filtration of samples $\{\tilde{z}_t\}_{t \ge 1}$ since the gradient is computed with an i.i.d. sample $\tilde{z}_t$,
    \begin{align*}
        \mathbb{E}[\xi_t | \mathcal{F}_{t-1}] &= \mathbb{E}[\nabla \ell_S(\gamma_{t-1}; \tilde{z}_t) | \mathcal{F}_{t-1}] - \nabla \widetilde{L}_S(\gamma_{t-1}) \\
        &= 0 \quad a.s.
    \end{align*}
    for all $t$. Furthermore, for all $t$
    \begin{align}
        \mathbb{E}[||\xi_t||_2^2 | \mathcal{F}_{t-1}] + ||\nabla \widetilde{L}_S(\gamma_{t-1})||_2^2 &\le \mathbb{E}[||\xi_t||_2^2 + ||\nabla \widetilde{L}_S(\gamma_{t-1})||_2^2 | \mathcal{F}_{t-1}] \nonumber \\
        &\le 2 \mathbb{E}[||\nabla \ell_S(\gamma_{t-1}; \tilde{z}_t)||_2^2 | \mathcal{F}_{t-1}] \nonumber \\
        &\le 2 C \cdot (1 + ||\gamma_{t-1}||_2^2) \label{eq:grad-noise-upper-bound}
    \end{align}
    where the last inequality follows from Assumption~\ref{assumption:glm-pop-loss-full}.
    
    Now consider the function $V(u) = ||u||_2^2$ which for all $u$, $z \in \mathbb{R}^d$ has a $2$-Lipschitz continuous gradient
    \begin{align*}
        ||\nabla V(u) - \nabla V(z)||_2 \le 2 \cdot ||u - v||_2.
    \end{align*}
    Let $\Delta_t = \gamma_t - \gamma^\ast$ and $\overline{R}(u) = \nabla\widetilde{L}_S(u + \gamma^\ast)$. Then the function $V_t = V(\Delta_t)$ after one update can be upper bounded using a second order Taylor expansion and the fact that $V$ has a Lipschitz-continuous gradient,
    \begin{align*}
        V_t &\le V_{t-1} + \nabla V_{t-1}^T (-\eta_t(\overline{R}(\Delta_{t-1}) + \xi_t)) + 2||\Delta_t - \Delta_{t-1}||_2^2 \\
        &\le V_{t-1} - \eta_t \nabla V_{t-1}^T \overline{R}(\Delta_{t-1}) - \eta_t \nabla V_{t-1}^T \xi_t + 2\eta_t^2 ||\overline{R}(\Delta_{t-1}) + \xi_t||_2^2.
    \end{align*}
    Taking the expectation conditioned on $\mathcal{F}_{t-1}$,
    \begin{align}
        \mathbb{E}[V_t | \mathcal{F}_{t-1}] &\le V_{t-1} - \eta_t \nabla V_{t-1}^T\overline{R}(\Delta_{t-1}) + 2 \eta_t^2 \mathbb{E}\left[||\overline{R}(\Delta_{t-1}) + \xi_t||_2^2 | \mathcal{F}_{t-1}\right] \nonumber \\
        &\le V_{t-1} - \eta_t \nabla V_{t-1}^T\overline{R}(\Delta_{t-1}) + 2 \eta_t^2 \bigg(||\overline{R}(\Delta_{t-1})||_2^2 + \mathbb{E}\left[||\xi_t||_2^2 | \mathcal{F}_{t-1}\right]\bigg) \nonumber \\
        &\le V_{t-1} - \eta_t \nabla V_{t-1}^T \overline{R}(\Delta_{t-1}) + 2C\eta_t^2(1 + ||\gamma_{t-1}||_2^2) \nonumber \\
        &\le V_{t-1}(1 + \eta_t^2 K) + \eta_t^2 K - \eta_t\nabla V_{t-1}^T \overline{R}(\Delta_{t-1}) \label{eq:rs-bound-pop}
    \end{align}
    where we use the fact that $\{\xi_t\}_{t\ge 1}$ is a martingale-difference sequence, the inequality from Eq. (\ref{eq:grad-noise-upper-bound}), and we gather constants into a suitable $K > 0$. Then we note that the terms $(1 + \eta_t^2 K)$, $\eta_t^2K \ge 0$ and by convexity of $\widetilde{L}_S$
    \begin{align*}
        \eta_t \nabla V_{t-1}^T \overline{R}(\Delta_{t-1}) &= 2\eta_t(\gamma_{t-1} - \gamma^\ast)^T \nabla \widetilde{L}_S(\gamma_{t-1}) \\
        &\ge 2\eta_t (\widetilde{L}(\gamma_{t-1}) - \widetilde{L}_S(\gamma^\ast)) \\
        &\ge 0.
    \end{align*}
    Then with the step size assumptions which ensure $\sum_{i=1}^\infty \eta_t = \infty$ and $\sum_{i=1}^\infty \eta_t^2 < \infty$ we obtain by the theorem of \citet{RS71} that $V_t \rightarrow V_\infty$ where $V_\infty$ is bounded.

    We now conclude this section by showing that $V_t$ in fact converges to 0. Begin by showing $\widetilde{L}_S$ is strongly convex in a neighborhood around $\gamma^\ast$. Consider $\gamma \in B(\gamma^\ast, r)$, the function is locally Lipschitz around $\gamma^\ast$,
    \begin{align*}
        \lambda_{min}(\nabla^2 \widetilde{L}_S(\gamma)) \ge m - \mathbb{E}[M(\tilde{z})] \cdot ||\gamma - \gamma^\ast||_2.
    \end{align*}
    Then taking the smaller of the two radii $r$ and $m/(2\mathbb{E}[M(\tilde{z})])$, for all $\gamma \in B(\gamma^\ast, \min\{r, m/(2\mathbb{E}[M(\tilde{z})])\}$ we have $\lambda_{min}(\nabla^2 \widetilde{L}_S(\gamma)) \ge m/2$. Continuing to work within this neighborhood,
    \begin{align*}
        (\gamma - \gamma^\ast)^T \nabla \widetilde{L}_S(\gamma) &= (\gamma - \gamma^\ast)^T \int_0^1 \nabla^2 \widetilde{L}_S(\gamma^\ast + t(\gamma - \gamma^\ast) ) (\gamma - \gamma^\ast) \ dt \\
        &\ge \frac{m}{2} \cdot ||\gamma - \gamma^\ast||_2^2.
    \end{align*}
    Now we showed that $V_t$ converges to $V_\infty$ which is bounded, thus, for any $\varepsilon > 0$, there exists $R$ such that $\Pr[\sup_t ||\Delta_t||_2 < R] \ge 1 - \varepsilon$. Define the stopping time $\tau_R = \inf \{t \ge 1 : ||\Delta_t||_2 \le R\}$, then on the event that $\hat{\gamma}$ is within radius $R$ of $\tau$, $\{t < \tau_R\}$ we have from Eq. (\ref{eq:rs-bound-pop}) and the bound above
    \begin{align*}
        \mathbb{E}[V_t I(t < \tau_R)] &\le \mathbb{E}\left[\left(1- \eta_t \frac{m}{2}+\eta_t^2 K\right) \cdot V_{t-1} I(t-1 < \tau_R)\right] + \eta_t^2 K.
    \end{align*}
    \citet{PJ92} then conclude that $\mathbb{E}[V_tI(t < \tau_R)]$ is upper bounded by the step size $\eta_t$ multiplied by a constant. As the step size converges to zero, this implies the almost sure convergence of the algorithm $||\Delta_t||_2^2 \rightarrow 0$.

    Now what remains is to prove that the distribution of error from the process $\{\Delta_t\}_{t \ge 1}$ is asymptotically normal. The argument by \citet{PJ92} proceeds by defining a process $\{\Delta_t'\}_{t\ge 1}$ which approximates the true sequence of errors $\{\Delta_t\}_{t\ge 1}$ with a linear update. Then by their Theorem 1, which applies to linear problems, the distribution of error in the proxy process is asymptotically normal, and it suffices to show that these two processes are asymptotically equivalent.
    
    For our setting of GLMs, define the proxy process
    \begin{align*}
        \Delta_{t}' &= \Delta_{t-1}' - \eta_t G\Delta_{t-1}' + \eta_t\xi_t, \qquad
        \Delta_0' = \Delta_0 \\
        \overline{\Delta}_{t} &= \frac{1}{t} \sum_{i=0}^{t-1} \Delta_i'
    \end{align*}
    where $G = \nabla^2 \widetilde{L}_S(\gamma^\ast)$. The matrix $G$ satisfies for any $\gamma \in B(\gamma^\ast, r)$, 
    \begin{align*}
        ||\nabla \widetilde{L}_S(\gamma) - G (\gamma-\gamma^\ast)||_2 &\le \left(\int_0^1 ||\nabla^2 \widetilde{L}_S(\gamma^\ast + t (\gamma - \gamma^\ast)) - \nabla^2 \widetilde{L}_S(\gamma^\ast)||_{op} \ dt\right) \cdot ||\gamma - \gamma^\ast||_2 \\
        &\le \mathbb{E}[M(\tilde{z})] \cdot ||\gamma - \gamma^\ast||_2^2
    \end{align*}
    which follows from the locally Lipschitz continuous Hessian around $\gamma^\ast$.
    
    The remaining assumptions for Theorem 2 are regarding the error decomposition $\xi_t = \xi_t(0) + \zeta_t(\Delta_{t-1})$ where
    \begin{align*}
        \xi_t(0) &= \nabla \ell_S(\gamma^\ast; \tilde{z}_t) - \nabla \widetilde{L}_S(\gamma^\ast) = \nabla \ell_S(\gamma^\ast; \tilde{z}_t) \\
        \zeta_t(\Delta_{t-1}) &= [\nabla \ell_S(\Delta_{t-1} + \gamma^\ast; \tilde{z}_t) - \nabla \ell_S(\gamma^\ast; \tilde{z}_t)] + [\nabla \widetilde{L}_S (\gamma^\ast) - \nabla \widetilde{L}_S(\Delta_{t-1} + \gamma^\ast)].
    \end{align*}
    that satisfy the following
    \begin{align}
        \mathbb{E}[\xi_t(0) | \mathcal{F}_{t-1}] &= 0 \quad a.s. \label{eq:error-cond-1}\\
        \mathbb{E}[\xi_t(0)\xi_t(0)^T | \mathcal{F}_{t-1}] &\overset{P}{\longrightarrow} S \quad \text{as } t \rightarrow \infty; \quad S > 0 \label{eq:error-cond-2}\\
        \sup_t \mathbb{E}[||\xi_t(0)||_2^2 I(||\xi_t(0)||_2 > C) | \mathcal{F}_{t-1}] &\overset{P}{\longrightarrow} 0  \quad \text{as } C \rightarrow \infty \label{eq:error-cond-3}\\
        \exists \ T, \forall t \ge T, \mathbb{E}[||\zeta_t(\Delta_{t-1})||_2^2 | \mathcal{F}_{t-1}] &\le \delta(\Delta_{t-1}) \quad a.s. \label{eq:error-cond-4}
    \end{align}
    with $\delta(\Delta_t) \rightarrow 0$ as $\Delta_t \rightarrow 0$. We now prove the properties (\ref{eq:error-cond-1}-\ref{eq:error-cond-4}). Updates are performed with an i.i.d. sample $\tilde{z}_t$, making the expectation of $\xi_t(0)$ conditioned on the previous samples zero and furthermore,
    \begin{align*}
        \mathbb{E}[\xi_t(0)\xi_t(0)^T | \mathcal{F}_{t-1}] &= \mathbb{E}[\nabla \ell_S(\gamma^\ast; \tilde{z}) \nabla \ell_S(\gamma^\ast; \tilde{z})^T] =: S,
    \end{align*}
    thus, meeting Properties (\ref{eq:error-cond-1}) and (\ref{eq:error-cond-2}). Now observe that for all $t$ conditional on previous samples, 
    \begin{align*}
        \mathbb{E}[||\xi_t(0)||_2^2 I(||\xi_t(0)||_2 > C) | \mathcal{F}_{t-1}] = \mathbb{E}[||\xi_1(0)||_2^2 I(||\xi_1(0)||_2 > C)].
    \end{align*}
    We know that $\mathbb{E}[||\xi_1(0)||_2^2] = \mathbb{E}[||\nabla \ell_S(\gamma^\ast; \tilde{z})||_2^2] < C \cdot (1 + ||\gamma^\ast||_2^2)$ by Assumption \ref{assumption:glm-pop-loss-full}. Now Property (\ref{eq:error-cond-3}) is equivalent to
    \begin{align*}
        \sup_t \mathbb{E}[||\xi_t(0)||_2^2 I (||\xi_t(0)||_2 > 0) | \mathcal{F}_{t-1}] &= \mathbb{E}[||\xi_1(0)||_2^2 I(||\xi_1(0)||_2 > C)]
    \end{align*}
    which converges to zero in probability as $C \rightarrow \infty$ since the expectation of $||\xi_1(0)||_2^2$ is finite. For Property (\ref{eq:error-cond-4}) note that we assume the gradients of $\ell_S$ are locally Lipschitz, thus,
    \begin{align*}
        \mathbb{E}[||\zeta_t(\Delta_{t-1})||_2^2 | \mathcal{F}_{t-1}] &\le \mathbb{E}[||\nabla \ell_S(\gamma_{t-1}; \tilde{z}_t) - \nabla \ell_S(\gamma^\ast; \tilde{z}_t)||_2^2 | \mathcal{F}_{t-1}] + ||\nabla \widetilde{L}_S(\gamma^\ast) - \nabla \widetilde{L}_S(\gamma_{t-1})||_2^2 \\
        &\le 2\mathbb{E}[M(\tilde{z})]^2 \cdot ||\Delta_{t-1}||_2^2
    \end{align*}
    which tends to zero once $\Delta_t \rightarrow 0$.
\end{proof}

\subsection{Proof of Theorem~\ref{theorem:glm-as-convergence-pop}}\label{appendix:glm-as-convergence}
\begin{proof}
    By Corollary~\ref{corollary:pj-for-pop-glms} the SGD sequence $\bar{\alpha}_t \rightarrow^{a.s.} \alpha^\ast$, likewise, $\bar{\beta}_t \rightarrow^{a.s.} \beta^\ast$. What remains is to show $\alpha^\ast = \beta^\ast$. By definition, the teacher estimator recovers the true parameter $\theta^\ast$. Consequently, the distillation label corresponding to a sample $(x, y)$ is $y' = h(x^T \theta^\ast) = \mathbb{E}[y| x]$. Both limit points $\alpha^\ast$ and $\beta^\ast$ must satisfy the first order optimality conditions for their respective objectives $L_S$ and $L_S'$ which correspond to the following conditions,
    \begin{align*}
        \mathbb{E}[h(v^T \alpha^\ast) v] = \mathbb{E}[yv], \quad \mathbb{E}[h(v^T \beta^\ast)v] = \mathbb{E}[y'v]. 
    \end{align*}
    The following equality $\mathbb{E}[yv] = \mathbb{E}[\mathbb{E}[y | x]v] = \mathbb{E}[y'v]$ from taking iterated expectations means that points $\alpha^\ast$ and $\beta^\ast$ must satisfy the same optimality condition.
    Then strict convexity implies that the optimal point is unique, thus showing that $\alpha^\ast$ and $\beta^\ast$ which satisfy the same condition for optimality must be the same point.
\end{proof}

\subsection{Proof of Theorem~\ref{theorem:glm-asymp-normality-pop}}
\begin{proof}
    Using Theorem~\ref{theorem:glm-as-convergence-pop}, the Hessians of the population losses evaluated at the optimal points are $H(\gamma^\ast) = \nabla^2 L_S(\gamma^\ast)$ and $H'(\gamma^\ast) = \nabla^2 L_S'(\gamma^\ast)$. Let $C(\gamma^\ast) = \mathbb{E}[\nabla \ell_S(\gamma^\ast; z) (\nabla \ell_S(\gamma^\ast; z))^T]$ and $C'(\gamma^\ast) = \mathbb{E}[\nabla \ell_S(\gamma^\ast; z') (\nabla \ell_S(\gamma^\ast; z'))^T]$ be the second moment of the gradient noise at the optimum $\gamma^\ast$. Then with Assumptions~\ref{assumption:glm-pop-loss-full} and \ref{assumption:pj-step-size}, we can apply Corollary~\ref{corollary:pj-for-pop-glms} which states that the averaged SGD iterates are asymptotically normal. Precisely,
    \begin{align*}
        \sqrt{t}(\bar{\alpha}_t - \gamma^\ast) \overset{d}{\longrightarrow} \mathcal{N}(0, H(\gamma^\ast)^{-1} C(\gamma^\ast) H(\gamma^\ast)^{-1}), \\
        \sqrt{t}(\bar{\beta}_t - \gamma^\ast) \overset{d}{\longrightarrow} \mathcal{N}(0, H'(\gamma^\ast)^{-1} C'(\gamma^\ast) H'(\gamma^\ast)^{-1}).
    \end{align*}
    The Hessians are equal,
    \begin{align*}
        H(\gamma^\ast) &= \mathbb{E}[h'(v^T\gamma^\ast) vv^T] = H'(\gamma^\ast)
    \end{align*}
    The expression for the gradients are $\nabla \ell_S(\gamma^\ast; z) = (h(v^T\gamma^\ast) - y)v$ and $\nabla \ell_S(\gamma^\ast; z') = (h(v^T \gamma^\ast) - y')v$. Let $\omega = h(v^T \gamma^\ast) - y'$ and $\delta = y' - y$. We can then write the gradient covariance $C(\gamma^\ast)$ as,
    \begin{align*}
        C(\gamma^\ast) &= \mathbb{E}[(h(v^T\gamma^\ast - y))^2 vv^T] \\
        &= \mathbb{E}[((h(v^T \gamma^\ast) - y') + (y' - y))^2 vv^T] \\
        &= \mathbb{E}[\omega^2 vv^T] + 2\mathbb{E}[\omega \delta vv^T] + \mathbb{E}[\delta^2 vv^T].
    \end{align*}
    The cross term is equal to zero, to see this take the conditional expectation
    \begin{align*}
        \mathbb{E}[\omega \delta vv^T] &= \mathbb{E}[\mathbb{E}[y' - y | x] \cdot (h(v^T\gamma^\ast) - y') vv^T] \\
        &= \mathbb{E}[\mathbb{E}[h(x^T \theta^\ast) - y) | x] \cdot (h(v^T\gamma^\ast) - h(x^T\theta^\ast)) vv^T]
    \end{align*}
    where for the last equality we simply write the definition of the distillation label. Making use of the fact that the teacher estimator is well-specified and thus $\mathbb{E}[y | x] = h(x^T \theta^\ast)$, the term above is equal to zero. The covariance is now $C(\gamma^\ast) = \mathbb{E}[\omega^2 vv^T] + \mathbb{E}[\delta^2 vv^T]$. The last expression has the familiar form $\mathbb{E}[\delta^2 vv^T] = \mathbb{E}[\text{Var}(y | x) vv^T]$. Now observe that the gradient covariance of the SGD sequence with knowledge distillation is,
    \begin{align*}
        C'(\gamma^\ast) &= \mathbb{E}[(h(v^T \gamma^\ast) - y')^2 vv^T].
    \end{align*}
    We can write $C(\gamma^\ast) = C'(\gamma^\ast) + \mathbb{E}[\text{Var}(y | x) vv^T]$, and it is now clear that $C(\gamma^\ast) \succeq C'(\gamma^\ast)$ since the variance component is always positive semi-definite.
\end{proof}

\section{Detailed assumptions and deferred proofs for Section~\ref{subsection:mle-finite-dataset-sgd}}\label{appendix:glm-sgd-finite}
\subsection{Assumptions}\label{appendix:glm-emp-assumptions}
We now present the formal assumptions in Theorems~\ref{appendix:glm-as-convergence-emp} and \ref{appendix:glm-asym-normality-emp}. All assumptions apply to both the standard supervision objectives, $\ell_S(\alpha; z)$ and $L_S^n(\alpha)$, as well as the knowledge distillation objectives, $\ell_S(\beta; z')$ and ${L'}_S^n(\beta)$. We unify the presentation by introducing the notation $\ell_S(\gamma; \tilde{z})$ and $\widetilde{L}_S^n(\gamma)$, with $\tilde{z} \in \{z, z'\}$. All assumptions are stated in this general form.

\begin{assumption}\label{assumption:glm-emp-loss-full}
The per-sample loss $\ell_S(\gamma; \tilde{z})$ and empirical loss $\widetilde{L}_S^n(\gamma) = 1/n \cdot \sum_{i=1}^n \ell_S(\gamma; \tilde{z}_i)$ where $\tilde{z} = (v, \tilde{y})$, $v=\Pi x$, and $\tilde{y} \in \{y, y'\}$ satisfies the following:
    \begin{enumerate}
        \item[(i)] At the unique optimal point $\hat{\gamma} := \argmin_\gamma \widetilde{L}_S^n(\gamma)$, for a constant $m > 0$, the Hessian $\nabla^2 \widetilde{L}_S^n(\hat{\gamma}) \succeq m \cdot I_d$ is positive definite.
        \item[(ii)] The map $\gamma \mapsto \ell_S(\gamma; \tilde{z})$ is twice continuously differentiable for every $z$. There exists constants $r$, $M > 0$ such that for all $\gamma \in B(\hat{\gamma}, r)$, 
        \begin{align*}
            ||\nabla \ell_S(\gamma; \tilde{z}) - \nabla \ell_S(\hat{\gamma}; \tilde{z})||_2 &\le M \cdot ||\gamma - \hat{\gamma}||_2 \\
            ||\nabla^2 \ell_S(\gamma; \tilde{z}) - \nabla^2 \ell_S(\hat{\gamma}; \tilde{z})||_{op} &\le M \cdot ||\gamma- \hat{\gamma}||_2.
        \end{align*}
        \item[(iii)] There exists a constant $C > 0$ such that the per-sample gradient $\nabla \ell_S(\gamma; \tilde{z})$, is bounded $||\nabla \ell_S(\gamma; \tilde{z})||_2^2 \le C \cdot (1 + ||\gamma||_2^2)$.
    \end{enumerate}
\end{assumption}

We now state Theorem 2 of \cite{PJ92}, specialized to the setting of GLMs with the empirical loss, as the following corollary.
\begin{corollary}[\citet{PJ92}, GLMs with objective $\widetilde{L}_S^n$] \label{corollary:pj-for-emp-glms}
    Consider the stochastic gradient descent update with sample $(v_t, \tilde{y}_t)$,
    \begin{align*}
        \gamma_t = \gamma_{t-1} - \eta_t \nabla \ell_S(\gamma_{t-1}; v_t, \tilde{y}_t).
    \end{align*}
    Assume \ref{assumption:glm-emp-loss-full} and \ref{assumption:pj-step-size} are true, then $\bar{\gamma}_t \rightarrow \hat{\gamma}$ almost surely, and
    \begin{align*}
        \sqrt{t} (\bar{\gamma}_t - \hat{\gamma}) \overset{d}{\longrightarrow} \mathcal{N}(0, \widetilde{H}_n^{-1}\widetilde{C}_n(\hat{\gamma}) \widetilde{H}_n^{-1})
    \end{align*}
    with second moment of the gradient noise at the optimum $\widetilde{C}_n(\hat{\gamma}) = 1/n \cdot \sum_{i=1}^n \nabla \ell_S(\hat{\gamma}; v_i, \tilde{y}_i) (\nabla \ell_S(\hat{\gamma}; v_i, \tilde{y}_i))^T$ and empirical Hessian $\widetilde{H}_n = 1/n \cdot \sum_{i=1}^n \nabla^2 \ell_S(\hat{\gamma}; v_i, \tilde{y}_i)$.
\end{corollary}

\begin{proof}
    We begin by writing the stochastic gradient descent update in the form analyzed in Theorem 2 of \citet{PJ92},
    \begin{align*}
        \gamma_t = \gamma_{t-1} - \eta_t g_t
    \end{align*}
    where $\xi_t = \nabla \ell_S(\gamma_{t-1}; v_t, \tilde{y}_t) - \nabla \widetilde{L}_S^n(\gamma_{t-1})$ and $g_t = \widetilde{L}_S^n(\gamma_{t-1}) + \xi_t$. Now we show that the assumptions of the original theorem hold, starting with the set of assumptions for proving almost sure convergence of SGD using the theorem by \citet{RS71}. Begin by noting that the gradient noise sequence $\{\xi_t\}_{t \ge 1}$ is a martingale-difference sequence with respect to the natural filtration of samples $\{\tilde{z}_t\}_{t \ge 1}$, for all $t$,
    \begin{align*}
        \mathbb{E}[\xi_t | \mathcal{F}_{t-1}] &= \mathbb{E}[\nabla \ell_S(\gamma_{t-1}; \tilde{z}_t) | \mathcal{F}_{t-1}] - \nabla \widetilde{L}_S^n(\gamma_{t-1}) \\
        &= 0 \quad a.s.
    \end{align*}
    where the expectation is taken over independent $\tilde{z}$ sampled uniformly from the finite dataset. Moreover, for all $t$
    \begin{align*}
        \mathbb{E}[||\xi_t||_2^2 | \mathcal{F}_{t-1}] + ||\nabla \tilde{L}_S^n(\gamma_{t-1})||_2^2 &= \frac{1}{n} \sum_{i=1}^n ||\nabla \ell_S(\gamma_{t-1}; \tilde{z}_i) - \nabla \widetilde{L}_S^n(\gamma_{t-1})||_2^2 + ||\nabla \widetilde{L}_S^n(\gamma_{t-1})||_2^2 \\
        &\le \frac{2}{n} \sum_{i=1}^n ||\nabla \ell_S(\gamma_{t-1}; \tilde{z}_i)||_2^2 \\
        &\le 2C \cdot (1 + ||\gamma_{t-1}||_2^2)
    \end{align*}
    where the final inequality follows from Assumption~\ref{assumption:glm-emp-loss-full}.

    Let $V(u) = ||u||_2^2$. This function has a 2-Lipschitz continuous gradient, for all $u$, $z \in \mathbb{R}^d$,
    \begin{align*}
        ||\nabla V(u) - \nabla V(z)||_2^2 \le 2 \cdot ||u - z||_2.
    \end{align*}
    Define the error $\Delta_t = \gamma_t - \hat{\gamma}$ and $\overline{R}(u) = \nabla \widetilde{L}_S^n(u + \hat{\gamma})$. Then we can derive an upper bound for $V_t = V(\Delta_t)$ by a second order Taylor expansion and the fact that $V$ has a 2-Lipschitz continuous gradient
    \begin{align*}
        V_t &\le V_{t-1} + \nabla V_{t-1}^T(-\eta_t(\overline{R}(\Delta_{t-1}) +\xi_t)) + 2||\Delta_t - \Delta_{t-1}||_2^2 \\
        &\le V_{t-1} - \eta_t \nabla V_{t-1}^T \overline{R}(\Delta_{t-1}) - \eta_t\nabla V_{t-1}^T\xi_t + 2\eta_t^2 ||\overline{R}(\Delta_{t-1}) + \xi_t||_2^2.
    \end{align*}
    Now taking expectations conditioned on the previous history, recall that the gradient noise is a martingale-difference sequence thus
    \begin{align}
        \mathbb{E}[V_t | \mathcal{F}_{t-1}] &\le V_{t-1} - \eta_t \nabla V_{t-1}^T \overline{R}(\Delta_{t-1}) + 2\eta_t^2 \left(||\overline{R}(\Delta_{t-1})||_2^2 + \mathbb{E}[||\xi_t||_2^2 | \mathcal{F}_{t-1}]\right) \nonumber \\
        &\le V_{t-1} - \eta_t \nabla V_{t-1}^T\overline{R}(\Delta_{t-1}) + 2C\eta_t^2(1 + ||\gamma_{t-1}||_2^2) \nonumber \\
        &\le V_{t-1}(1 + \eta_t^2 K) + \eta_t^2 K -\eta_t \nabla V_{t-1}^T \overline{R}(\Delta_{t-1}) \label{eq:rs-bound-emp}
    \end{align}
    where the second to last inequality follows from Assumption~\ref{assumption:glm-emp-loss-full} and we gather constants into a suitable $K > 0$. To apply the theorem by \citet{RS71} which proves $V_t$ converges to a bounded random variable, we must show these coefficients are nonnegative. This is clear as $(1+\eta_t^2K)$, $\eta_t^2K \ge 0$ and by convexity of $\widetilde{L}_S^n$ we have
    \begin{align*}
        \eta_t \nabla V_{t-1}^T \overline{R}(\Delta_{t-1}) &= 2\eta_t (\gamma_{t-1} - \hat{\gamma})^T \nabla \widetilde{L}_S^n(\gamma_{t-1}) \\
        &\ge 2\eta_t(\widetilde{L}_S^n(\gamma_{t-1}) - \widetilde{L}_S^n(\hat{\gamma})) \\
        &\ge 0
    \end{align*}
    since $\hat{\gamma}$ minimizes $\widetilde{L}_S^n$. With our step size assumptions $\sum_{i=1}^\infty \eta_t = \infty$ and $\sum_{i=1}^\infty \eta_t^2 < \infty$ we can apply the theorem by \citet{RS71} to get that $V_t \rightarrow V_\infty$ where $V_\infty$ is bounded. 

    We now conclude this section of this proof by showing $V_t$ actually converges to 0. The idea is to show that the bound in Eq. (\ref{eq:rs-bound-emp}) contracts $V_t$. We show that $\widetilde{L}_S^n$ is strongly convex in a neighborhood around $\hat{\gamma}$. For all $\gamma \in B(\hat{\gamma}, r)$ the smallest eigenvalue of the Hessian is,
    \begin{align*}
        \lambda_{min}(\nabla \widetilde{L}_S^n) \ge m - M\cdot ||\gamma - \hat{\gamma}||_2.
    \end{align*}
    Define the ball with the minimum of the two radii $r$ and $m/(2M)$, and the Hessian at points in this set is bounded below by $m/2 \cdot I_d$. It follows that for all $\gamma \in B(\hat{\gamma}, \min\{r, m/(2M)\})$,
    \begin{align*}
        (\gamma - \hat{\gamma})^T\nabla \widetilde{L}_S^n(\gamma) &= (\gamma - \hat{\gamma})^T \int_{0}^1 \nabla^2 \widetilde{L}_S^n(\hat{\gamma} + t(\gamma - \hat{\gamma})) (\gamma - \hat{\gamma}) \ dt \\
        &\ge \frac{m}{2} \cdot ||\gamma-\hat{\gamma}||_2^2.
    \end{align*}
    Earlier we proved that $V_t$ converges to $V_\infty$ which is bounded, therefore, for any $\varepsilon > 0$, there exists $R$ such that $\Pr[\sup_t ||\Delta_t||_2 < R] \ge 1-\varepsilon$. Let $\tau_R = \inf\{t \ge 1 : ||\Delta_t||_2 \le R\}$ be the stopping time. Then on $\{t < \tau_R\}$ we have from Eq. (\ref{eq:rs-bound-emp}) and bound above
    \begin{align*}
        \mathbb{E}[V_t I(t < \tau_R)] \le \mathbb{E}\left[\left(1 - \eta_t \frac{m}{2} + \eta_t^2 K\right) \cdot V_{t-1} I(t-1 < \tau_R)\right] + \eta_t^2 K.
    \end{align*}
    The proof by \citet{PJ92} concludes that $\mathbb{E}[V_t I(t < \tau_R)]$ is upper bounded by $K\cdot \eta_t$. Thus, as the step size converges to zero, $V_t$ converges to zero, the error $||\Delta_t||_2^2 \rightarrow 0$.

    In the second part of the proof we show that the distribution of error from the process $\{\Delta_t\}_{t\ge1}$ is asymptotically normal. \cite{PJ92} prove this by defining a proxy process $\{\Delta_t'\}_{t\ge1}$ which is a linear approximation of the true process $\{\Delta_t\}_{t\ge1}$. Then they apply their Theorem 1 to $\Delta_t'$, which shows that the distribution of errors is asymptotically normal. Lastly, they conclude with showing that the two process are asymptotically equivalent.

    For our setting, the proxy process is defined as
    \begin{align*}
        \Delta_t' &= \Delta_{t-1}' - \eta_t G_n \Delta_{t-1}' + \eta_t\xi_t, \quad \Delta_0' = \Delta_0 \\
        \overline{\Delta}_t &= \frac{1}{t}\sum_{i=0}^{t-1}\Delta_i'
    \end{align*}
    where $G_n = \nabla^2 \widetilde{L}_S^n(\hat{\gamma})$. Then using the fact that the locally Lipschitz-continuous Hessian around $\hat{\gamma}$, the matrix satisfies for all $\gamma \in B(\hat{\gamma}, r)$,
    \begin{align*}
        ||\nabla \widetilde{L}_S^n(\gamma) - G_n(\gamma - \hat{\gamma})||_2 &\le \left(\int_0^1 ||\nabla^2 \widetilde{L}_S^n(\hat{\gamma} + t(\gamma - \hat{\gamma})) - \nabla^2 \widetilde{L}_S^n(\hat{\gamma})||_{op} \ dt\right) \cdot ||\gamma - \hat{\gamma}||_2 \\
        &\le M \cdot ||\gamma - \hat{\gamma}||_2^2.
    \end{align*}

    The final assumptions used in the proof of Theorem 2 revolve around the error decomposition $\xi_t = \xi_t(0) + \zeta_t(\Delta_{t-1})$ where
    \begin{align*}
        \xi_t(0) &= \nabla \ell_S(\hat{\gamma}; \tilde{z}_t) - \nabla\widetilde{L}_S^n(\hat{\gamma}) \\
        \zeta_t(\gamma_{t-1}) &= [\nabla\ell_S(\gamma_{t-1}; \tilde{z}_t) - \nabla \ell_S(\hat{\gamma}; \tilde{z}_t)] + [\nabla \widetilde{L}_S^n(\hat{\gamma}) - \nabla \widetilde{L}_S^n(\gamma_{t-1})].
    \end{align*}
    which satisfy the following properties
    \begin{align}
        \mathbb{E}[\xi_t(0) | \mathcal{F}_{t-1}] &= 0 \quad a.s. \label{eq:emp-error-cond-1}\\
        \mathbb{E}[\xi_t(0)\xi_t(0)^T | \mathcal{F}_{t-1}] &\overset{P}{\longrightarrow} S \quad \text{as } t \rightarrow \infty; \quad S > 0 \label{eq:emp-error-cond-2}\\
        \sup_t \mathbb{E}[||\xi_t(0)||_2^2 I(||\xi_t(0)||_2 > C) | \mathcal{F}_{t-1}] &\overset{P}{\longrightarrow} 0  \quad \text{as } C \rightarrow \infty \label{eq:emp-error-cond-3}\\
        \exists \ T, \forall t \ge T, \mathbb{E}[||\zeta_t(\Delta_{t-1})||_2^2 | \mathcal{F}_{t-1}] &\le \delta(\Delta_{t-1}) \quad a.s. \label{eq:emp-error-cond-4}
    \end{align}
    with $\delta(\Delta_t) \rightarrow 0$ as $\Delta_t \rightarrow 0$. Given that each update is performed with $\tilde{z}_t$ sampled uniformly from the finite dataset, we have
    \begin{align*}
        \mathbb{E}[\xi_t(0)|\mathcal{F}_{t-1}] &= \mathbb{E}[\nabla \ell_S(\hat{\gamma}; \tilde{z})] - \nabla \widetilde{L}_S^n(\hat{\gamma}) \\
        &= 0 \quad a.s.
    \end{align*}
    Furthermore, conditioned on the previous samples
    \begin{align*}
        \mathbb{E}[\xi_t(0)\xi_t(0)^T | \mathcal{F}_{t-1}] = \frac{1}{n} \sum_{i=1}^n\nabla \ell_S(\hat{\gamma}; \tilde{z}_i) (\nabla \ell_S(\hat{\gamma}; \tilde{z}_i))^T =: S
    \end{align*}
    which proves Properties (\ref{eq:emp-error-cond-1}) and (\ref{eq:emp-error-cond-2}). Again, using the fact that updates are performed with independent samples, for all $t$ 
    \begin{align*}
        \mathbb{E}[||\xi_t(0)||_2^2 I(||\xi_t(0)||_2 > C) | \mathcal{F}_{t-1}] = \mathbb{E}[||\xi_t(0)||_2^2 I(||\xi_1(0)||_2 > C)].
    \end{align*}
    By Assumption~\ref{assumption:glm-emp-loss-full} the expectation is finite $\mathbb{E}[||\xi_1(0)||_2^2] < C \cdot (1+||\hat{\gamma}||_2^2)$. This turns Property (\ref{eq:emp-error-cond-3}) into the equivalent statement
    \begin{align*}
        \sup_t \mathbb{E}[||\xi_t(0)||_2^2 I(||\xi_t(0)||_2 > C) | \mathcal{F}_{t-1}] = \mathbb{E}[||\xi_1(0)||_2^2 I(||\xi_1(0)||_2 > C)]
    \end{align*}
    which converges to zero in probability as $C \rightarrow \infty$. Lastly,
    \begin{align*}
        \mathbb{E}[||\zeta_t(\gamma_{t-1})||_2^2 | \mathcal{F}_{t-1}] &\le \mathbb{E}[||\nabla \ell_S(\gamma_{t-1}; \tilde{z}_t) - \nabla \ell_S(\hat{\gamma}; \tilde{z}_t)||_2^2 | \mathcal{F}_{t-1}] + ||\nabla \widetilde{L}_S^n(\hat{\gamma}) - \nabla \widetilde{L}_S^n(\gamma_{t-1})||_2^2 \\
        &\le 2M^2 \cdot ||\gamma - \hat{\gamma}||_2^2
    \end{align*}
    which converges to zero once $\Delta_t \rightarrow 0$, thus, proving Property (\ref{eq:emp-error-cond-4}).
\end{proof}

We also use the following assumptions to ensure concentration occurs.
\begin{assumption}\label{assumption:emp-concentration-full}
    The covariates and labels are such that:
    \begin{enumerate}
        \item[(i)] There exists a constant $\mu > 0$ such that the matrix, 
        \begin{align*}
            \mathbb{E}[\text{Var}(y | x) vv^T] \succeq \mu I_d.
        \end{align*}
        \item[(ii)] There exists constants $r$, $b > 0$ such that for all $\gamma \in B(\gamma^\ast, r)$,
        \begin{align}
            \sup_{\gamma \in B} ||\nabla \ell_S(\gamma; \tilde{z})||_2 \le b.
        \end{align}
    \end{enumerate}
\end{assumption}

\subsection{Proof of Theorem~\ref{theorem:glm-as-convergence-emp}}\label{appendix:glm-as-convergence-emp}
\begin{proof}
    By Corollary~\ref{corollary:pj-for-emp-glms}, the sequences $\bar{\alpha}_t \rightarrow^{a.s.} \hat{\alpha}$ and $\bar{\beta}_t \rightarrow^{a.s.} \hat{\beta}$. This convergence ensures that the subsequent analysis on $\hat{\alpha}$ and $\hat{\beta}$ are valid. Now what's left is to show that $\hat{\alpha} = \hat{\beta}$. The gradient of the per-sample loss $\nabla \ell_T(\theta; x, y) = (h(x^T\theta) - y)x$. Since the teacher estimator $\hat{\theta}$ minimizes the empirical loss, we have
    \begin{align}
        \frac{1}{n} \sum_{i=1}^n y_i x_i = \frac{1}{n} \sum_{i=1}^n h(x_i^T \hat{\theta}) x_i. \label{eq:emp-teacher-first-order-eq}
    \end{align}
    The gradient of the per-sample loss of the student is $\nabla \ell_S(\gamma; v, y) = (h(v^T\gamma) - y)v$. Both points $\hat{\alpha}$ and $\hat{\beta}$ satisfies the first-order optimality conditions involving their respective labels, which corresponds to the following equations,
    \begin{align*}
        \frac{1}{n} \sum_{i=1}^n y_i v_i &= \frac{1}{n} \sum_{i=1}^n h(v_i^T \hat{\alpha}) v_i, \\
        \frac{1}{n} \sum_{i=1}^n h(x_i^T \hat{\theta}) v_i &= \frac{1}{n} \sum_{i=1}^n h(v_i^T \hat{\beta}) v_i.
    \end{align*}
    The objective $L_S^n$ is strictly convex and thus has a unique optimal point. Furthermore, Eq.~\ref{eq:emp-teacher-first-order-eq} shows both $\hat{\alpha}$ and $\hat{\beta}$ satisfy the first-order optimality conditions on either labels. This implies the points are the same.
\end{proof}

\subsection{Proof of Theorem~\ref{theorem:glm-asym-normality-emp}}\label{appendix:glm-asym-normality-emp}
\begin{proof}
    Recall that both methods share an optimal point by Theorem~\ref{theorem:glm-as-convergence-emp}. The Hessians at that optimal point $\hat{\gamma}$ are defined as $H_n(\hat{\gamma}) = 1/n \cdot \sum_{i=1}^n \nabla^2 \ell_S(\hat{\gamma}; z_i)$ and $H_n'(\hat{\gamma}) = 1/n \cdot \sum_{i=1}^n \nabla^2 \ell_S(\hat{\gamma}; z_i')$. The empirical second moments of the gradient noise at $\hat{\gamma}$ are,
    \begin{align*}
        C_n(\hat{\gamma}) &= \frac{1}{n} \cdot \sum_{i=1}^n \nabla \ell_S(\hat{\gamma}; z_i) (\nabla \ell_S(\hat{\gamma}; z_i))^T \\
        C_n'(\hat{\gamma}) &= \frac{1}{n} \cdot \sum_{i=1}^n \nabla \ell_S(\hat{\gamma}; z_i') (\nabla \ell_S(\hat{\gamma}; z_i'))^T.
    \end{align*}
    Corollary~\ref{corollary:pj-for-emp-glms} states under Assumptions~\ref{assumption:glm-emp-loss-full} and \ref{assumption:pj-step-size}, that the averaged iterates are asymptotically normal,
    \begin{align*}
        \sqrt{t}(\bar{\alpha}_t - \hat{\gamma}) \overset{d}{\longrightarrow} \mathcal{N}(0, [H_n(\hat{\gamma})]^{-1} C_n(\hat{\gamma}) [H_n(\hat{\gamma})]^{-1}) \\
        \sqrt{t}(\bar{\beta}_t - \hat{\gamma}) \overset{d}{\longrightarrow} \mathcal{N}(0, [H_n'(\hat{\gamma})]^{-1} C_n'(\hat{\gamma}) [H_n'(\hat{\gamma})]^{-1}).
    \end{align*}
    Observe that the Hessians are equal,
    \begin{align*}
        H_n(\hat{\gamma}) = \frac{1}{n} \sum_{i=1}^n h'(v_i^T\hat{\gamma}) v_iv_i^T = H_n'(\hat{\gamma}),
    \end{align*}
    going forward we refer to both of them as $H_n$. The theorem is now equivalent to showing $C_n(\hat{\gamma}) \succeq C_n'(\hat{\gamma})$. At a high level, we go about proving this is by bounding the deviation of the empirical quantities, namely the parameter estimates as well as the empirical second moments, from their corresponding population quantities. 
    
    To show the difference in covariances is positive semi-definite, we want to bound the minimum eigenvalue of $C_n(\hat{\gamma}) - C_n'(\hat{\gamma})$ and show that it is nonnegative. We know that their population counterpart $C(\gamma^\ast) - C'(\gamma^\ast) = \mathbb{E}[\text{Var}(y | x) vv^T]$ is positive definite under Assumption~\ref{assumption:emp-concentration}. For any symmetric matrices $A$, $B \in \mathbb{R}^{d\times d}$ we can lower bound the sum of one's eigenvalues by $\lambda_{min}(A) \ge \lambda_{min}(B) + \lambda_{min}(A-B)$. The difference $A - B$ is still symmetric, which means $\lambda_{min}(A-B) \ge -\lambda_{max}(|A-B|) = -||A-B||_{op}$. We apply this identity to write the difference between the empirical second moments evaluated at the minimizers of the empirical objective, in terms of the deviations from their respective population counterparts,
    \begin{align*}
        \lambda_{min} \bigg[C_n(\hat{\gamma}) - C_n'(\hat{\gamma}) \bigg] \ge \lambda_{min} \bigg[C(\gamma^\ast) - C'(\gamma^\ast)\bigg] - ||C_n(\hat{\gamma}) - C(\gamma^\ast)||_{op}- ||C_n'(\hat{\gamma}) - C'(\gamma^\ast)||_{op}.
    \end{align*}
    Note that the minimum eigenvalue of $C(\gamma^\ast) - C'(\gamma^\ast)$ is at least $\mu > 0$. Let $R = \min\{r, m/(2M)\}$ where $r$ is the radius from our assumptions regarding the function's local Lipschitz properties. The function $L_S^n$ is strongly convex around $\hat{\gamma}$, consider $\gamma \in B(\hat{\gamma}, R)$ then
    \begin{align*}
        \lambda_{min}(\nabla^2 L_S^n(\gamma)) &\ge m - ||\nabla^2 L_S^n(\hat{\gamma}) - \nabla^2 L_S^n(\gamma)||_{op} \\
        &\ge m - M\cdot ||\hat{\gamma} - \gamma||_2 \\
        &\ge m/2.
    \end{align*}
    Suppose $\gamma \not\in B(\hat{\gamma}, R)$ and let $d = ||\gamma - \hat{\gamma}||_2$, then
    \begin{align*}
        (\gamma - \hat{\gamma})^T \nabla L_S^n(\gamma) &= (\gamma - \hat{\gamma})^T \int_0^1 \nabla^2 L_S^n(\hat{\gamma} + t(\gamma - \hat{\gamma})) (\gamma - \hat{\gamma}) \ dt \\
        &\ge (\gamma - \hat{\gamma})^T \int_0^{R/d} \frac{m}{2} (\gamma - \hat{\gamma}) \ dt \\
        &\ge \frac{mR}{2d} ||\gamma - \hat{\gamma}||_2^2.
    \end{align*}
    Then by Cauchy-Schwarz the gradient norm at $\gamma$ is $||\nabla L_S^n(\gamma)||_2 \ge mR/2$. This implies that if $||\nabla L_S^n(\gamma)||_2 < mR/2$, then $\gamma \in B(\hat{\gamma}, R)$. We show for sufficiently large $n$, the gradient norm at $\gamma^\ast$ is less than $mR/2$. We use Assumption~\ref{assumption:glm-emp-loss-full} to bound each coordinate of $\nabla L_S^n(\gamma^\ast)$ via Hoeffding's inequality, we get for all $t \ge 0$
    \begin{align*}
        \Pr[|(\nabla L_S^n(\gamma^\ast))_i| \ge t] \le 2\exp\left(-\frac{nt^2}{2b^2}\right).
    \end{align*}
    Then a union bound over all coordinates gives us, with probability $\ge 1-\delta$,
    \begin{align*}
        ||\nabla L_S^n(\gamma^\ast)||_\infty \le b \sqrt{\frac{2\log(2d/\delta)}{n}}.
    \end{align*}
    And the $\ell_2$ bound follows, $||\nabla L_S^n(\gamma^\ast)||_2 \le \sqrt{d} \cdot ||\nabla L_S^n(\gamma^\ast)||_\infty$. We then take $n$ large enough so that the gradient norm is less than $mR/2$, which implies $\gamma^\ast \in B(\hat{\gamma}, R)$. Consider the term involving $C_n(\hat{\gamma}) - C(\gamma^\ast)$,
    \begin{align*}
        ||C_n(\hat{\gamma}) - C(\gamma^\ast)||_{op} &\le ||C_n(\hat{\gamma}) - C_n(\gamma^\ast)||_{op} + ||C_n(\gamma^\ast) - C(\gamma^\ast)||_{op}.
    \end{align*}
    By Assumptions~\ref{assumption:glm-emp-loss-full} and \ref{assumption:emp-concentration-full}, we know that the per-sample gradients are locally Lipschitz and bounded in $B(\gamma^\ast, r)$, it follows that
    \begin{align*}
        ||C_n(\hat{\gamma}) - C_n(\gamma^\ast)||_{op} &\le \frac{1}{n} \sum_{i=1}^n ||\nabla \ell_S(\hat{\gamma}; z_i) (\nabla \ell_S(\hat{\gamma}; z_i))^T - \nabla\ell_S(\gamma^\ast; z_i)(\nabla\ell_S(\gamma^\ast; z_i))^T||_{op} \\
        &\le \frac{1}{n}\sum_{i=1}^n ||\nabla \ell_S(\hat{\gamma}; z) - \nabla \ell_S(\gamma^\ast; z)||_2 (||\nabla \ell_S(\hat{\gamma}; z)||_2 + ||\nabla \ell_S(\gamma^\ast; z)||_2) \\
        &\le 2M\cdot b ||\hat{\gamma} - \gamma^\ast||_2.
    \end{align*}
    Now we derive a bound on the second term using the matrix Bernstein inequality~\citep{T15}. For sample $(x_i, y_i) \sim \mathcal{P}$ where $z_i = (v_i, y_i)$ and $v_i = \Pi x_i$, let
    \begin{align*}
        M_i := \nabla\ell_S(\gamma^\ast; z_i)(\nabla\ell_S(\gamma^\ast; z_i))^T - C(\gamma^\ast).
    \end{align*}
    Then $\mathbb{E}[M_i] = 0$ by definition. Under boundedness (Assumption~\ref{assumption:emp-concentration-full}),
    \begin{align*}
        ||C(\gamma^\ast)||_{op} &\le \mathbb{E}[||\nabla \ell_S(\gamma^\ast; z) (\nabla \ell_S(\gamma^\ast; z))^T||_{op}] \\
        &= \mathbb{E}[||\nabla \ell_S(\gamma^\ast; z)||_2^2] \le b^2.
    \end{align*}
    We then have an upper bound on the variance proxy via the triangle inequality,
    \begin{align*}
        ||\mathbb{E}[M_i^2]||_{op} &\le \mathbb{E}[||M_i||_{op}^2] \\
        &\le \mathbb{E}[(||\nabla\ell_S(\gamma^\ast; z)||_2^2 + ||C(\gamma^\ast)||_{op})^2] \\
        &\le 4b^4.
    \end{align*}
    Using these quantities along with the matrix Bernstein inequality, for all $t \ge 0$
    \begin{align*}
        \Pr\left[||C_n(\gamma^\ast) - C(\gamma^\ast)||_{op} \ge t\right] &=  \Pr\left[\bigg|\bigg|\frac{1}{n} \cdot \sum_{i=1}^n M_i\bigg|\bigg|_{op} \ge t\right] \\
        &\le 2d \cdot \exp\left(-\frac{nt^2/2}{4b^4 + 2b^2t/3}\right).
    \end{align*}
    So with probability $\ge 1-\delta$, we have
    \begin{align*}
        ||C_n(\gamma^\ast) - C(\gamma^\ast)||_{op} \le b^2 \left(\sqrt{\frac{8\cdot \log(2d /\delta)}{n}} + \frac{2\cdot\log(2d/\delta)}{3n}\right).
    \end{align*}
    A sufficiently large $n$ means that $||C_n(\hat{\gamma}) - C(\gamma^\ast)||_{op} \le \varepsilon'$ with high probability. Given the same assumptions but for $\ell_S(\gamma; z')$, ${L'}_S$, and ${L'}_S^n$, the prior argument applies verbatim. Thus,
    \begin{align*}
        ||C_n'(\hat{\gamma}) - C'(\gamma^\ast)||_{op} &\le ||C_n'(\hat{\gamma}) - C_n'(\gamma^\ast)||_{op} + ||C_n'(\gamma^\ast) - C'(\gamma^\ast)||_{op} \\
        &\le \varepsilon'
    \end{align*}
    with high probability. Let $\varepsilon' = \mu/4$. We then have with high probability,
    \begin{align*}
        ||C_n(\hat{\gamma}) - C(\gamma^\ast)||_{op} + ||C_n'(\hat{\gamma}) - C'(\gamma^\ast)||_{op} \le \mu/2
    \end{align*}
    which means the minimum eigenvalue of $C_n(\hat{\gamma}) - C_n'(\hat{\gamma})$ is at least $\mu/2$. Lastly, we can conclude that for sufficiently large $n$,
    \begin{align*}
        H_n^{-1}C_n(\hat{\gamma}) H_n^{-1} \succeq H_n^{-1}C_n'(\hat{\gamma}) H_n^{-1}
    \end{align*}
    holds with high probability.
\end{proof}

\section{Detailed statements and deferred proofs for Section~\ref{section:low-rank}}\label{appendix:rank-1}

We introduce the notation $\widetilde{M} \in \{M, M'\}$ to simplify the presentation of our results when the argument for either target matrix is identical.

To begin, we may assume that $\widetilde{M}$ is diagonal. To see this, consider the eigendecomposition $\widetilde{M} = U \Lambda U^T$. Since $\widetilde{M}$ is symmetric, $U$ is an orthogonal matrix. For the rotated iterates $y_t = U^Tx_t$, the update rule is
\begin{align*}
    y_{t+1} &= y_t - \eta(y_ty_t^T - \Lambda)y_t.
\end{align*}
Thus, the dynamics are equivalent up to a rotation, so it is no loss of generality to work in the eigenbasis of $\widetilde{M}$. Now let's make some general observations about the gradient descent dynamics.

\begin{lemma}[Bounded magnitude]\label{lemma:rank1-bounded-magnitude}
    Under initialization $x_0 \sim \mathcal{N}(0, k\cdot \lambda_1/n \cdot I_n)$ and for step size $\eta < 1/(L \cdot (\lambda_1 + 1))$ where $0 < k < 1$ is a chosen constant and $L \ge 4$, elements of the gradient descent sequence given by the update:
    \begin{align*}
        x_{t+1} = x_t - \eta (x_tx_t^T - \widetilde{M})x_t,
    \end{align*}
    have squared norm most $\lambda_1$ with probability $1-k$ over the initialization.
\end{lemma}
\begin{proof}
    We first consider the event that the squared norm at initialization is less than $\lambda_1$. By Markov's inequality we have $\Pr[||x_0||_2^2 \ge \lambda_1] \le \mathbb{E}[||x_0||_2^2] / \lambda_1 = k$ since $\mathbb{E}[||x_0||_2^2] = k \cdot \lambda_1$. It follows that $\Pr[||x_0||_2^2 < \lambda_1] > 1- k$. 
    
    We have the following gradient update for the squared norm:
    \begin{align*}
        ||x_{t+1}||_2^2 &= \sum_{i=1}^n (1 - \eta(||x_t||_2^2 - \lambda_i))^2 x_{i, t}^2 \\
        &\le (1 - \eta(||x_t||_2^2 - \lambda_1))^2 ||x_t||_2^2
    \end{align*}
    where the last inequality follows from taking the update with the largest possible eigenvalue, which is $\lambda_1$. Note that by choice of $\eta$ we have $(1-\eta(||x_t||_2^2 - \lambda_i)) \ge 0$ for all $i = 1, \ldots, n$. Now we proceed via induction conditioned on the event that the initial squared norm is less than $\lambda_1$. Since the base case is satisfied, moving to the inductive step, assume $||x_t||_2^2 \le \lambda_1$. Consider the following function $f$ and its derivative,
    \begin{align*}
        f(z) &= (1-\eta(z - \lambda_1))^2 z, \\
        f'(z) &= (1-\eta(z - \lambda_1))^2 -2\eta(1-\eta(z - \lambda_1))z \\    
        &=(1-\eta(z - \lambda_1))(1-\eta(3z - \lambda_1)) \ge 0
    \end{align*}
    The derivative is positive for all $z \in [0, \lambda_1]$ since $\eta < 1/(L \cdot (\lambda_1 + 1))$. It follows that in this interval, $f(z) \le f(\lambda_1) = \lambda_1$. From the inductive hypothesis, $||x_{t+1}||_2^2 \le f(||x_t||_2^2) \le \lambda_1$, so we can conclude that $||x_t||_2^2 \le \lambda_1$ for all $t \ge 1$.
\end{proof}

\subsection{Proof of Theorem~\ref{theorem:rank-1-convergence-to-mins}}\label{appendix:rank-1-convergence-to-mins}

\begin{proof}
    The convergence of GD sequences $w_t$ and $u_t$ to their respective local minima is a result of Theorem 3 in \citet{PP17}, which relaxes the globally Lipschitz assumption required in \cite{LSJR16}. This is necessary for the simple reason that $\nabla L_S$ and $\nabla L_S'$ are not globally Lipschitz. To demonstrate this for $\nabla L_S$ consider $x_1 = (\kappa + 1) \cdot v$ and $x_2 = \kappa \cdot v$ where $v \in \mathbb{R}^n$ has unit norm. Clearly, $||x_1 - x_2||_2 = 1$ so now derive,
    \begin{align*}
        \nabla L_S(x_1) - \nabla L_S(x_2) &= (||x_1||_2^2 I - M)x_1 - (||x_2||_2^2 I - M)x_2 \\
        &= ((\kappa + 1)^2 I - M) (\kappa + 1) \cdot v - (\kappa^2 I - M) \kappa \cdot v \\
        &= (3 \kappa^2 + 3\kappa + 1)\cdot v - Mv,
    \end{align*}
    then by the reverse triangle inequality we have,
    \begin{align*}
        ||\nabla L_S(x_1) - \nabla L_S(x_2)||_2 \ge |(3 \kappa^2 + 3\kappa + 1) - ||Mv||_2|.
    \end{align*}
    The term $||Mv||_2$ is a constant and $\kappa$ can be made arbitrarily large, making it impossible for $||\nabla L_S(x_1) - \nabla L_S(x_2)||_2 \le L||x_1-x_2||_2$ for a constant $L \ge 0$.
    
    The functions $\nabla L_S$ and $\nabla L_S'$ are locally Lipschitz on the open ball of radius $\sqrt{\lambda_1 + \varepsilon}$ for a small constant $0 < \varepsilon < 1$, 
    \begin{align*}
        \mathcal{B} = \{x : ||x||_2 < \sqrt{\lambda_1 + \varepsilon}\}.
    \end{align*}
    This set is forward invariant under the two maps $g(x) = x - \eta \nabla L_S(x)$ and $g'(x) = x - \eta \nabla L_S'(x)$ because we've assumed that at initialization, $||x_0||_2 \le \sqrt{\lambda_1}$, so directly applying Lemma~\ref{lemma:rank1-bounded-magnitude} allows us to conclude $g(\mathcal{B}) \subseteq \mathcal{B}$ and $g'(\mathcal{B}) \subseteq \mathcal{B}$. On this set,
    \begin{align*}
        \sup_{x \in \mathcal{B}} ||\nabla^2 L_S(x)||_{op} &= \sup_{x \in \mathcal{B}} ||2xx^T + ||x||_2^2 \cdot I - M||_{op} \\
        &\le \sup_{x \in \mathcal{B}} ||2xx^T + ||x||_2^2 \cdot I||_{op} + ||M||_{op} \\
        &\le 4\lambda_1 + 3 < \infty.
    \end{align*}
    In the last inequality, we use the fact that $x \in \mathcal{B}$ implies $||x||_2 < \sqrt{\lambda_1 + 1}$. Since the target matrix in $L_S'$ is the truncated eigendecomposition of $M$ to the top-$r$ eigenvalues, $||M'||_{op} = \lambda_1$. Consequently, the supremum of the operator norm of $\nabla^2 L'_S(x)$ over this set is still upper bounded by $4 \lambda_1 + 3$. Then by Theorem 3 in \citet{PP17}, the set of initial conditions $x \in \mathcal{B}$ so that either of the gradient descent trajectories with step size $0 < \eta < 1/(4\lambda_1 + 3)$ converges to a strict saddle point is of measure zero.
    
    Now if we derive the gradient of our objectives, $\nabla L_S(x) = (xx^T - M)x$ and $\nabla L_S'(x) = (xx^T - M')x$. It follows that the set of all critical points is $\{\pm \sqrt{\lambda_i} e_i\}_{i=1}^n \cup \{0\}$ for $L_S$ and $\{\pm \sqrt{\lambda_i}e_i\}_{i=1}^r \cup \{0\}$ for $L_S'$. All points except for $\sqrt{\lambda_1} e_1$ and $-\sqrt{\lambda_1}e_1$ are strict saddles. To see this, compute the Hessians and evaluate them at each critical point,
    \begin{align*}
        \nabla^2 L_S(\pm \sqrt{\lambda_i}e_i) &= \lambda_i \cdot I + 2 \lambda_i e_ie_i^T - M \\
        \nabla^2 L_S'(x) &= \lambda_i \cdot I + 2\lambda_ie_ie_i^T - M'.
    \end{align*}
    Along the direction $e_1$, they are $e_1^T \nabla^2 L_S(\pm \sqrt{\lambda_i}e_i) e_1 = (\lambda_i -\lambda_1) < 0$ and $e_1^T \nabla^2 L_S'(\pm \sqrt{\lambda_i}e_i) e_1 = (\lambda_i -\lambda_1) < 0$ when $i \ne 1$. Also note that the Hessians at 0 are simply $-M$ and $-M'$ respectively, and both have a negative eigenvalue $-\lambda_1$. Thus, all critical points except $\sqrt{\lambda_1} e_1$ and $-\sqrt{\lambda_1}e_1$ are strict saddles, and consequently, gradient descent avoids them. At $x=\pm\sqrt{\lambda_1}e_1$, the Hessians are both positive semi-definite, so these points are local minima (and also global minima since $\lambda_1 > \lambda_2$) of both objectives.
\end{proof}

\subsection{Full statement and proof of Lemma~\ref{lemma:rank1-lower-bound}}\label{appendix:rank1-lower-bound}

\begin{lemma}[Supervised learning lower bound]\label{lemma:rank1-lower-bound-full}
    Given $\varepsilon > 0$, let the step size be
    \begin{align*}
        0 < \eta < \frac{1}{L\cdot (\lambda_1 + 1)}
    \end{align*}
    where $\lambda_1$ is the largest eigenvalue of target matrix $M$ and $L \ge 4$. Assume the initial point $w_0$ satisfies $||w_0||_2 \le \sqrt{\lambda_1}$. Then $T(w)$ has the following lower bound,
    \begin{align*}
        T(w) \ge \frac{1}{2\log(1 / (1-\eta\Delta_n))} \log\left(\frac{(1 - \varepsilon) \cdot ||w_{2:n, 0}||_2^2 / w_{1, 0}^2}{\varepsilon}\right).
    \end{align*}
\end{lemma}
\begin{proof}[Proof of Lemma~\ref{lemma:rank1-lower-bound-full}]
    We begin by introducing some notation. For a vector $w_t$, its orthogonal projection onto the $i$-th eigenvector of $M$, $v_i$, is $P_{v_i} w_t$ where $P_{v_i} = v_iv_i^T$. As we're working with a diagonal $M$, $P_{v_i} w_t = w_{i, t}$ and we have the following one step update:
    \begin{align}\label{eq:rank1-component-update}
        w_{i, t+1} = (1 - \eta(||w_t||_2^2 - \lambda_i)) \cdot w_{i, t}.
    \end{align}
    In general, any vector $w_t$ can be decomposed as,
    \begin{align*}
        w_t = P_{v_1} w_t + P_{v_2, \ldots, v_n} w_t
    \end{align*}
    where $P_{v_1}$ and $P_{v_2, \ldots, v_n}$ are orthogonal projections onto the eigenvector(s) $v_1$ and $\{v_2, \ldots, v_n\}$ respectively. Now we can define the following quantity:
    \begin{align*}
        c_t &= \frac{||P_{v_2, \ldots, v_n} w_t||_2}{||P_{v_1} w_t||_2} = \frac{||w_{2:n, t}||_2}{|w_{1, t}|}
    \end{align*}
    where the last equality is again from working with a diagonal $M$. This value measures the tangent of the angle between $w_t$ (projected onto the eigenvectors orthogonal to $v_1$) and $w^\ast$. When the direction of $w_t$ is perfectly aligned with $w^\ast$ it is zero. Let's now consider the sequence $\{c_t^2\}_{t\ge1}$, it satisfies
    \begin{align*}
        c_{t+1}^2 &\ge \left(\frac{1 - \eta(||w_t||_2^2 - \lambda_n)}{1- \eta(||w_t||_2^2 - \lambda_1)}\right)^2\cdot c_t^2,
    \end{align*}
    where the lower bound is from Eq.~\ref{eq:rank1-component-update} with the smallest eigenvalue $\lambda_n$. Adding and subtracting by $\eta\lambda_1$ in the numerator decouples the overall magnitude from the update,
    \begin{align}
        c_{t+1}^2 &\ge \left(1 - \eta\frac{\lambda_1 - \lambda_n}{1-\eta(||w_t||_2^2 - \lambda_1)}\right)^2 \cdot c_t^2 \\
        &\ge (1-\eta\Delta_n)^2 \cdot c_t^2 \label{eq:lower-bound-angle-update}
    \end{align}
    where we define $\Delta_n := \lambda_1 - \lambda_n$. The lower bound uses $||w_t||_2^2 \le \lambda_1$ for all $t\ge 1$ from Lemma~\ref{lemma:rank1-bounded-magnitude}, which means the fraction $1/(1-\eta(||w_t||_2^2 - \lambda_1)) \le 1$. 

    We now connect the quantity $c_t$ to the main goal of lower bounding $T(w)$ for a given tolerance $\varepsilon > 0$. Observe that $||w_{2:n, t}||_2^2 = w_{1, t}^2 a_t^2$. As the eigenvectors of $M$ are orthogonal (here they are also the basis vectors $e_1, \ldots, e_n$), the distance from $w_t$ to $w^\ast$ can be written as,
    \begin{align*}
        ||w_t - w^\ast||_2^2 &= (\sqrt{\lambda_1} - w_{1, t})^2 + ||w_{2:n, t}||_2^2 \\
        &= (\sqrt{\lambda_1} - w_{1, t})^2 + w_{1, t}^2 c_t^2.
    \end{align*}
    Consider function,
    \begin{align*}
        g(z) = (\sqrt{\lambda_1} - z)^2 + z^2\cdot c_t^2,
    \end{align*}
    treating $c_t^2$ as fixed. It's strictly convex, so by taking the derivative $g'(z) = 2z\cdot c_t^2 - 2(\sqrt{\lambda_1} - z)$ we find that the function is minimized at $z = \sqrt{\lambda_1} / (1 + c_t^2)$. From this we get the lower bound,
    \begin{align*}
        ||w_t - w^\ast||_2^2 \ge \lambda_1 \frac{c_t^2}{1 + c_t^2}.
    \end{align*}
    The distance from the optimum can therefore be estimated from just angular information. By Eq.~\ref{eq:lower-bound-angle-update}, iterating the bound shows that $c_t^2 \ge (1 - \eta\Delta_n)^{2t} \cdot c_0^2$. The quantity $c_t^2$ must be at least $\varepsilon / (1 - \varepsilon)$ in order for $||w_t - w^\ast||_2^2$ to be at least $\lambda_1 \cdot \varepsilon$. Solving for $t$ at the desired value of $c_t$ yields a lower bound on the steps,
    \begin{align*}
        T(w) \ge \frac{1}{2\log(1/(1 - \eta\Delta_n))} \log\left(\frac{(1-\varepsilon) \cdot c_0^2}{\varepsilon}\right).
    \end{align*}
    Note that the notation used in the main body of this paper is $\textsc{Im}(w_0) = c_0^2.$
\end{proof}

\subsection{Full statement and proof of Lemma~\ref{lemma:rank1-upper-bound}}\label{appendix:rank1-upper-bound}
\begin{lemma}[Knowledge distillation upper bound]\label{lemma:rank1-upper-bound-full}
    Given $\varepsilon > 0$, let the step size be
    \begin{align*}
        0 < \eta < \frac{1}{L\cdot (\lambda_1 + 1)}
    \end{align*}
    where $\lambda_1$ is the largest eigenvalue of target matrix $M'$ and $L \ge 4$. Assume the initial point $u_0$ satisfies $||u_0||_2 \le \sqrt{\lambda_1}$. Let $\theta, \xi, \gamma \in (0, 1)$ be fixed constants, and define
    \begin{align*}
        \ell := \frac{L}{L+1}, \quad \kappa := \frac{1-\xi}{1+\gamma\varepsilon}.
    \end{align*}
    Define the following quantities,
    \begin{align*}
        t_a &:= \frac{1}{2\log(1/(1-\ell\eta\Delta_2))} \log\left(\frac{||u_{2:r, 0}||_2^2/u_{1, 0}^2}{\varepsilon_a}\right),\\
        t_b &:= \frac{1}{2\log(1/(1-\ell\eta\lambda_1))} \log\left(\frac{||u_{r+1:n, 0}||_2^2 / u_{1, 0}^2}{\varepsilon_b}\right),\\
        t_\xi &:= \frac{1}{2\log(1+\eta\xi\lambda_1)} \log\left(\frac{(1-\xi)\lambda_1}{u_{1, 0}^2}\right),\\ 
        t_\delta &:= \frac{1}{\log(1/(1-2\eta\kappa\lambda_1))}\log\left(\frac{1-\kappa}{\varepsilon_\delta - \kappa^{-1}(\varepsilon_a + \varepsilon_b)}\right).
    \end{align*}
    Here, the error tolerances are given by
    \begin{align*}
        \varepsilon_a := (\gamma\theta)\cdot\varepsilon, \quad \varepsilon_b := (1-\theta)\gamma \cdot \varepsilon, \quad \varepsilon_\delta := (1-\gamma)\cdot \varepsilon .
    \end{align*}
    Then $T(u)$ has the following upper bound
    \begin{align*}
        T(u) &\le \max\{t_a, t_b, t_\xi\} + t_\delta.
    \end{align*}
\end{lemma}

\begin{proof}[Proof of Lemma~\ref{lemma:rank1-upper-bound-full}]
    Similar to the proof of the lower bound, we begin by observing that any vector $u_t$ has the decomposition,
    \begin{align*}
        u_t = P_{v_1} u_t + P_{v_2, \ldots, v_r} u_t + P_{v_{r+1}, \ldots, v_n} u_t
    \end{align*}
    where $P_{v_1}$, $P_{v_2, \ldots, v_r}$, and $P_{v_{r+1}, \ldots, v_n}$ are orthogonal projections onto the eigenvector(s) $v_1$, $\{v_2, \ldots, v_r\}$, and $\{v_{r+1}, \ldots, v_n\}$ respectively. Note that $v_{r+1}, \ldots, v_n$ correspond to the eigenvalue 0 and span the nullspace. Now we define the following quantities:
    \begin{align*}
        a_t &= \frac{||P_{v_2, \ldots, v_r} u_t||_2}{||P_{v_1} u_t||_2} = \frac{||u_{2:r, t}||_2}{|u_{1, t}|}, \\
        b_t &= \frac{||P_{v_{r+1}, \ldots, v_n} u_t||_2}{||P_{v_1} u_t||_2} = \frac{||u_{r+1:n, t}||_2}{|u_{1, t}|}
    \end{align*}
    which represent tangent of the angle between $u_t$, projected onto the different subspaces, and $u^\ast$. The reason we introduce these two quantities is that the distance $||u_t - u^\ast||_2^2$ can be upper bounded by two monotonically decreasing sequences,
    \begin{align*}
        ||u_t - u^\ast||_2^2 &= (\sqrt{\lambda_1} - u_{1, t})^2 + ||u_{2:n, t}||_2^2 \\
        &= (\sqrt{\lambda_1} - u_{1, t})^2 + u_{1, t}^2 (a_t^2 + b_t^2) \\
        &\le (\lambda_1 - u_{1, t}^2) + \lambda_1 (a_t^2 + b_t^2).
    \end{align*}
    The last inequality follows from the identity $(\sqrt{\lambda_1} - u_{1, t})^2 \le (\sqrt{\lambda_1} - |u_{1, t}|)(\sqrt{\lambda_1} + |u_{1, t}|)$. Additionally, Lemma~\ref{lemma:rank1-bounded-magnitude} shows the magnitude $||u_t||_2^2 \le \lambda_1$ is bounded for all $t \ge 1$ with high probability over the initialization, the first coordinate $u_{1, t}$ also inherits this bound. We can upper bound $\{a_t^2\}_{t\ge1}$ with a monotonically decreasing sequence,
    \begin{align}
        a_{t+1}^2 &\le \left(\frac{1-\eta(||u_t||_2^2 - \lambda_2)}{1-\eta(||u_t||_2^2 - \lambda_1)}\right)^2 \cdot a_t^2 \\
        &= \left(1 - \eta \frac{\lambda_1 - \lambda_2}{1 - \eta(||u_t||_2^2 - \lambda_1)}\right)^2 \cdot a_t^2 \\
        &\le (1-\eta\ell\Delta_2)^2 \cdot a_t^2 \label{eq:rank1-upper-bound-a}
    \end{align}
    where we define $\Delta_2 := \lambda_1 - \lambda_2$. The constant $\ell := L/(L+1)$ appears due to the upper bound $(1- \eta(||u_t||_2^2 - \lambda_1)) \le 1+ \eta\lambda_1 \le (L+1)/L$ since the step size is chosen as $\eta < 1/(L \cdot (\lambda_1 + 1))$ and we can drop the additional 1 in the denominator. A similar argument shows the sequence $\{b_t\}_{t\ge1}$ satisfies,
    \begin{align}
        b_{t+1}^2 &= \left(\frac{1-\eta||u_t||_2^2}{1 - \eta(||u_t||_2^2 - \lambda_1)}\right)^2 \cdot b_t^2 \\
        &= \left(1 - \eta\frac{\lambda_1}{1 - \eta(||u_t||_2^2 - \lambda_1)}\right)^2 \cdot b_t^2 \\
        &\le (1 - \eta\ell \lambda_1)^2 \cdot b_t^2. \label{eq:rank1-upper-bound-b}
    \end{align}
    The key detail is that the rate depends on the top eigenvalue $\lambda_1$ instead of the spectral gap $\Delta_i$. Consequently, the $b_t$ terms (potentially) decay at a much faster rate when the magnitude of the eigenvalues is large, but the spectrum is tightly clustered.
    
    Let $\varepsilon_a$, $\varepsilon_b > 0$ be given tolerances. Solve for the number of steps such that $a_t \le \varepsilon_a$ using Eq.~\ref{eq:rank1-upper-bound-a},
    \begin{align*}
        t_a &:= \min\{t: a_t^2 \le \varepsilon_a\} \\
        &\le \frac{1}{2\log(1/(1-\eta\ell \Delta_2))} \log \left(\frac{a_0^2}{\varepsilon_a}\right).
    \end{align*}
    Likewise, using Eq.~\ref{eq:rank1-upper-bound-b} the number of steps such that $b_t \le \varepsilon_b$ is
    \begin{align*}
        t_b &= \min\{t : b_t^2 \le \varepsilon_b\} \\
        &\le \frac{1}{2\log(1 / (1-\eta\ell\lambda_1))} \log \left(\frac{b_0^2}{\varepsilon_b}\right).
    \end{align*}
    
    Now we analyze the growth of the first coordinate $u_{1, t}$. It has the following update equation,
    \begin{align}\label{eq:rank1-first-coordinate-update}
        u_{1, t+1} = (1-\eta(||u_t||_2^2 -\lambda_1))u_{1, t}.
    \end{align}
    The analysis consists of two phases, namely, the initial growth of the squared norm $||u_t||_2^2$ to $(1-\xi)\lambda_1$ and then a final convergence stage of $u_{1, t}^2$ to $\lambda_1$. For the latter, we specifically derive an update for $\delta_t := \lambda_1 - u_{1, t}^2$ which corresponds to the error in magnitude at step $t$. Starting with the full magnitude, while $||u_t||_2^2 < (1-\xi)\lambda_1$ the update on the full magnitude is lower bounded by the update on the first coordinate,
    \begin{align*}
        ||u_{t+1}||_2^2 &\ge (1+\eta\xi\lambda_1)^2 u_{1, t}^2.
    \end{align*}
    It follows that $||u_t||_2^2 \ge (1-\eta\xi\lambda_1)^{2t}u_{1, 0}^2$. Then solving for $t$ leads to the upper bound on steps to reach $(1-\xi)\lambda_1$,
    \begin{align*}
        t_\xi &:= \min\{t : ||u_t||_2^2 \ge (1-\xi)\lambda_1\} \\
        &\le \frac{1}{2\log(1+\eta\xi\lambda_1)} \log\left(\frac{(1-\xi)\lambda_1}{u_{1, 0}^2}\right).
    \end{align*}
    Now we analyze the final convergence of the first coordinate $u_{1, t}^2$ to target magnitude $\lambda_1$, this requires deriving a one-step update for $\delta_t$. At a high level, we know $\delta_t$ is monotonically decreasing because the magnitude $||u_t||_2^2$ is at most $\lambda_1$. Simply inspecting Eq.~\ref{eq:rank1-first-coordinate-update}, shows that the coefficient $(1 - \eta(||u_t||_2^2 - \lambda_1))$ is always nonnegative. In fact, it is always positive unless the magnitude is exactly $\lambda_1$, which means $u_{1, t}^2$ is increasing and $\delta_t$ is decreasing. To derive the actual update equation, observe that the sum of the magnitude gap at time $t$, $\delta_t$, and the growth of the first coordinate from step $t$ to $t+1$ is exactly $\delta_{t+1}$,
    \begin{align}
        \delta_{t+1} &= (\lambda_1 - u_{1, t}^2) + (u_{1, t}^2 - u_{1, t+1}^2) \nonumber\\
        &= \delta_t + u_{1, t}^2 - u_{1, t+1}^2. \label{eq:rank1-delta-expression}
    \end{align}
    Now we can focus on the growth captured by $u_{1, t}^2 - u_{1, t+1}^2$. Using Eq.~\ref{eq:rank1-first-coordinate-update}, the coordinate at $t+1$ can be written in terms of $u_{1, t}$ as follows,
    \begin{align*}
        u_{1, t}^2 - u_{1, t+1}^2 &= u_{1, t}^2 - (1-\eta(||u_t||_2^2 - \lambda_1))^2 \cdot u_{1, t}^2 \\
        &= u_{1, t}^2 - \bigg[1 - 2\eta (||u_t||_2^2 - \lambda_1) + \eta^2(||u_t||_2^2 - \lambda_1)^2\bigg] \cdot u_{1, t}^2 \\
        &= \bigg[2\eta (||u_t||_2^2 - \lambda_1) - \eta^2(||u_t||_2^2 - \lambda_1)^2\bigg] \cdot u_{1, t}^2 \\
        &\le 2\eta(||u_t||_2^2 - \lambda_1) \cdot u_{1, t}^2.
    \end{align*}
    From Lemma~\ref{lemma:rank1-bounded-magnitude}, we know that the magnitude $||u_t||_2^2 \le \lambda_1$ is bounded which makes the term involving $\eta^2$ nonpositive. Dropping it results in the final inequality. Now write the expansion,
    \begin{align*}
        2\eta(||u_t||_2^2 - \lambda_1) \cdot u_{1, t}^2 &= 2\eta(u_{1, t}^2 - \lambda_1) u_{1, t}^2 + 2\eta u_{1, t}^2 ||u_{2:n, t}||_2^2 \\
        &= -2\eta u_{1,t}^2 \cdot \delta_t  + 2\eta u_{1, t}^2 ||u_{2:n, t}||_2^2 \\
        &= -2\eta u_{1, t}^2 \cdot \delta_t + 2 \eta u_{1, t}^4 (a_t^2 + b_t^2)
    \end{align*}
    where the last equality uses the identity $||u_{2:n, t}||_2^2 = u_{1, t}^2 (a_t^2 + b_t^2)$ for the angular quantities defined earlier. Lastly, $2 \eta u_{1, t}^4 (a_t^2 + b_t^2) \le 2\eta \lambda_1^2 (a_t^2 + b_t^2)$ so that the bounding sequence is cleanly decreasing. Substituting this into Eq.~\ref{eq:rank1-delta-expression}, we have
    \begin{align*}
        \delta_{t+1} \le (1-2\eta u_{1, t}^2) \delta_t + 2\eta \lambda_1^2 (a_t^2 + b_t^2).
    \end{align*}
    This is the final convergence stage so we want to upper bound the number of additional steps, after $\max\{t_a, t_b, t_\xi\}$, in order for $\delta_t \le \lambda_1 \cdot \varepsilon_\delta$ for a given tolerance $\varepsilon_\delta > 0$. To be precise, 
    \begin{align*}
        t_\delta := \min\{t \ge \max\{t_a, t_b, t_\xi\} : \delta_t \le \lambda_1 \cdot \varepsilon_\delta\}.
    \end{align*}
    After $T_{1} := \max\{t_a, t_b, t_\xi\}$, the top coordinate $u_{1, t}^2$ occupies most of the magnitude. This is from the identity $u_{1, t}^2 = ||u_t||_2^2/(1 + a_t^2 + b_t^2)$. For $t \ge \max\{t_a, t_b, t_\xi\}$, we've already proven that $a_t^2 + b_t^2 \le \varepsilon_a + \varepsilon_b$ as well as $||u_t||_2^2 \ge (1-\xi)\cdot \lambda_1$. To simplify notation, let $\kappa := (1-\xi)/ (1 + \varepsilon_a + \varepsilon_b)$. The first coordinate is then at least $u_{1, t}^2 \ge \kappa \cdot \lambda_1$ for all $t \ge T_1$ and the difference at time $T_1$ is $\delta_{T_1} \le (1-\kappa)\cdot\lambda_1$. Now we can write an upper bound for $\delta_t$ that holds when $t \ge \max\{t_a, t_b, t_\xi\}$,
    \begin{align*}
        \delta_{t+1} &\le (1-2\eta\kappa \lambda_1)\delta_t + 2\eta\lambda_1^2 (a_t^2 + b_t^2) \\
        &\le (1-2\eta\kappa\lambda_1)^{t'+1}\delta_{T_1} + 2\eta\lambda_1^2 \cdot (\varepsilon_a + \varepsilon_b) \sum_{i=0}^{t'} (1-2\eta\kappa\lambda_1)^i
    \end{align*}
    where $t' := t-T_1$. For the last inequality we unroll the recursion and use the fact that $(a_t^2 + b_t^2)$ is nonincreasing and has the value $(a_{T_1}^2 + b_{T_1}^2) \le \varepsilon_a + \varepsilon_b$ at step $T_1$. Replacing the finite sum with an infinite sum to remove the dependence on $t'$, we arrive at the final upper bound
    \begin{align*}
        \delta_t &\le (1-2\eta\kappa\lambda_1)^{t'} \delta_{T_1} + 2\eta\lambda_1^2 \cdot (\varepsilon_a + \varepsilon_b) \cdot \frac{1}{1 - (1 - 2\eta\kappa\lambda_1)} \\
        &= (1-2\eta\kappa\lambda_1)^{t'} \delta_{T_1} + \frac{\lambda_1 \cdot (\varepsilon_a + \varepsilon_b)}{\kappa},
    \end{align*}
    from which we get the step upper bound,
    \begin{align*}
        t_\delta \le \frac{1}{\log(1/(1-2\eta\kappa\lambda_1))} \log\left(\frac{1-\kappa}{\varepsilon_\delta - (\varepsilon_a + \varepsilon_b)/\kappa}\right).
    \end{align*}
    Then at step $t = \max\{t_a, t_b, t_\xi\} + t_\delta$, the following is true,
    \begin{align*}
        a_t^2 + b_t^2 &\le \varepsilon_a + \varepsilon_b\\
        \delta_t &\le \lambda_1 \cdot \varepsilon_\delta,
    \end{align*}
    and consequently,
    \begin{align*}
        ||u_t - u^\ast||_2^2 &\le \lambda_1 \cdot (\varepsilon_a + \varepsilon_b + \varepsilon_\delta).
    \end{align*}
    Lastly, $\varepsilon_a = (\gamma\theta)\cdot \varepsilon$, $\varepsilon_b = (1-\theta)\gamma \cdot \varepsilon$, and $\varepsilon_\delta = (1-\gamma) \cdot \varepsilon$ for constants $\theta$, $\xi$, $\gamma \in (0, 1)$. Therefore, the distance is bounded $||u_t - u^\ast||_2^2 \le\ \lambda_1 \cdot \varepsilon$. The overall upper bound is then $T(u) \le \max\{t_a, t_b, t_\xi\} + t_\delta$.
\end{proof}

\subsection{Proof of Theorem~\ref{theorem:conditions-for-kd-gd-faster}}


\begin{proof}[Proof of Theorem~\ref{theorem:conditions-for-kd-gd-faster}]
    From Lemmas~\ref{lemma:rank1-lower-bound-full} and \ref{lemma:rank1-upper-bound-full}, we have a lower bound on $T(w)$,
    \begin{align*}
        T(w) \ge \frac{1}{2\log(1/(1 - \eta\Delta_n))} \log\left(\frac{(1-\varepsilon) \cdot c_0^2}{\varepsilon}\right)
    \end{align*}
    and an upper bound on $T(u)$ (taking $\xi =1/2$),
    \begin{align*}
        T(u) \le \max\{t_a, t_b, t_\xi\} + t_\delta.
    \end{align*}
    The quantities $T(w)$ and $T(u)$ correspond to the number of steps needed to reach a given error tolerance by supervised learning and knowledge distillation, respectively. We want to understand conditions on the original matrix $M$, such that the difference $T(w) - T(u)$ is positive. Recall that we initialize $n_0 \sim \mathcal{N}(0, k\cdot\lambda_1/n \cdot I_n)$ with some $k \in (0, 1)$ and set $w_0 = u_0 = n_0$. For each realization of $n_0$, define
    \begin{align*}
        k_a = \frac{r}{n} ||n_{2:r, 0}||_2^2, \quad k_b = \frac{(n-r)}{n} ||n_{r+1:n, 0}||_2^2, \quad k_1 = \frac{1}{n} n_{1, 0}^2.
    \end{align*}
    We previously defined
    \begin{align*}
        a_t = \frac{||u_{2:r, t}||_2}{|u_{1, t}|}, \quad b_t = \frac{||u_{r+1:n, t}||_2}{|u_{1, t}|},\quad c_t = (a_t^2 + b_t^2)^{1/2}.
    \end{align*}
    At initialization we have $a_0^2 = k_a/k_1 \cdot r$ and $b_0^2 = k_b/k_1 \cdot (n-r)$. Also note that by the concentration of the chi-square random variable~\citep{LM00}, one can identify fixed intervals such that $a_0^2$ and $b_0^2$ fall within these intervals with constant probability. The endpoints of these intervals replace the constants $k_a$, $k_b$, and $k_1$ that appear as the multiplicative factors of $\varepsilon^{-c}$ in the matrix dimension lower bound. Importantly, this substitution does not change the exponent $c$ nor does it affect the conclusion that $n=\Omega(\varepsilon^{-c})$. Now define the following quantities
    \begin{align*}
        d_w &:= 2\log(1/(1-\eta\Delta_n)) \\
        d_a &:= 2\log(1/(1-\ell\eta\Delta_2)) \\
        d_b &:= 2\log(1/(1-\ell\eta\lambda_1)) \\
        d_\xi &= 2\log(1+\eta\xi\lambda_1) \\
        d_\delta &:= \log(1/(1-2\eta\kappa\lambda_1)).
    \end{align*}
    
    We consider the different cases of the maximum. The first case is when $t_a \ge t_b, t_\xi$, then the difference is,
    \begin{align*}
        T(w) - T(u) &\ge \frac{d_a d_\delta - d_w(d_a+d_\delta)}{d_a d_\delta d_w} \log\left(\frac{1}{\varepsilon}\right) + \frac{\log(n)}{d_w} - \frac{\log(r)}{d_a} + R_a
    \end{align*}
    where the remainder $R_a$ does not depend on matrix dimension $n$,
    \begin{align*}
        R_a = \frac{\log((k_a + k_b)/ k_1)}{d_w} - \frac{\log(k_a/ (k_1\theta \cdot \gamma))}{d_a} - \frac{1}{d_\delta}\log\left(\frac{1}{1-\gamma(1+\kappa^{-1})}\right).
    \end{align*}
    This leads to the following constraint on the dimension $n$,
    \begin{align*}
        n \ge \exp\left(\frac{d_w(d_a+d_\delta) - d_ad_\delta}{d_ad_\delta}\log\left(\frac{1}{\varepsilon}\right) + \frac{d_w}{d_a}\log(r) - d_wR_a\right).
    \end{align*}
    The second case is when $t_b \ge t_a, t_\xi$.
    \begin{align*}
        T(w) - T(u) \ge \frac{d_bd_\delta - d_w(d_b+d_\delta)}{d_bd_\delta d_w}\log\left(\frac{1}{\varepsilon}\right) + \frac{d_b - d_w}{d_bd_w}\log(n) + R_b
    \end{align*}
    where the $n$-independent constant is
    \begin{align*}
        R_b = \frac{\log((k_a + k_b)/k_1)}{d_w} - \frac{\log(k_b / (k_1(1-\theta) \cdot \gamma))}{d_b} - \frac{1}{d_\delta}\log\left(\frac{1}{1-\gamma(1+\kappa^{-1})}\right).
    \end{align*}
    By assumption $d_b - d_w > 0$, thus the dimension $n$ must then be at least,
    \begin{align*}
        n \ge \exp\left(\frac{d_w(d_b + d_\delta) - d_bd_\delta}{d_\delta(d_b - d_w)}\log\left(\frac{1}{\varepsilon}\right) - R_b\frac{d_bd_w}{d_b-d_w}\right).
    \end{align*}
    The final case is $t_\xi \ge t_a, t_b$,
    \begin{align*}
        T(w) - T(u) \ge \frac{d_\delta - d_w}{d_\delta d_w}\log\left(\frac{1}{\varepsilon}\right) + \frac{d_\xi - d_w}{d_\xi d_w}\log (n) + R_\xi
    \end{align*}
    with dimension-independent constant
    \begin{align*}
        R_\xi = \frac{\log((k_a + k_b)/k_1)}{d_w} - \frac{\log(1/k_1)}{d_\xi} - \frac{1}{d_\delta}\log\left(\frac{1}{1-\gamma(1+\kappa^{-1})}\right).
    \end{align*}
    As we've assumed $d_\xi - d_w > 0$, it follows that the dimension $n$ must be,
    \begin{align*}
        n \ge \exp\left(\frac{d_\xi(d_w-d_\delta)}{d_\delta(d_\xi-d_w)}\log\left(\frac{1}{\varepsilon}\right) - R_\xi \frac{d_\xi d_w}{d_\xi - d_w}\right).
    \end{align*}
    Now we can conclude by taking,
    \begin{align*}
        c = \max\left\{\frac{d_w(d_a+d_\delta) - d_ad_\delta}{d_ad_\delta}, \frac{d_w(d_b + d_\delta) - d_bd_\delta}{d_\delta(d_b - d_w)}, \frac{d_\xi(d_w-d_\delta)}{d_\delta(d_\xi-d_w)}\right\},
    \end{align*}
    and required matrix dimension is $\Omega(\varepsilon^{-c})$.
\end{proof}

\clearpage
\section{Additional experiments and details}
\subsection{Logistic regression on synthetic data}\label{appendix:glm-synthetic}
\begin{figure}[h!]
    \centering
    \includegraphics[width=0.9\linewidth]{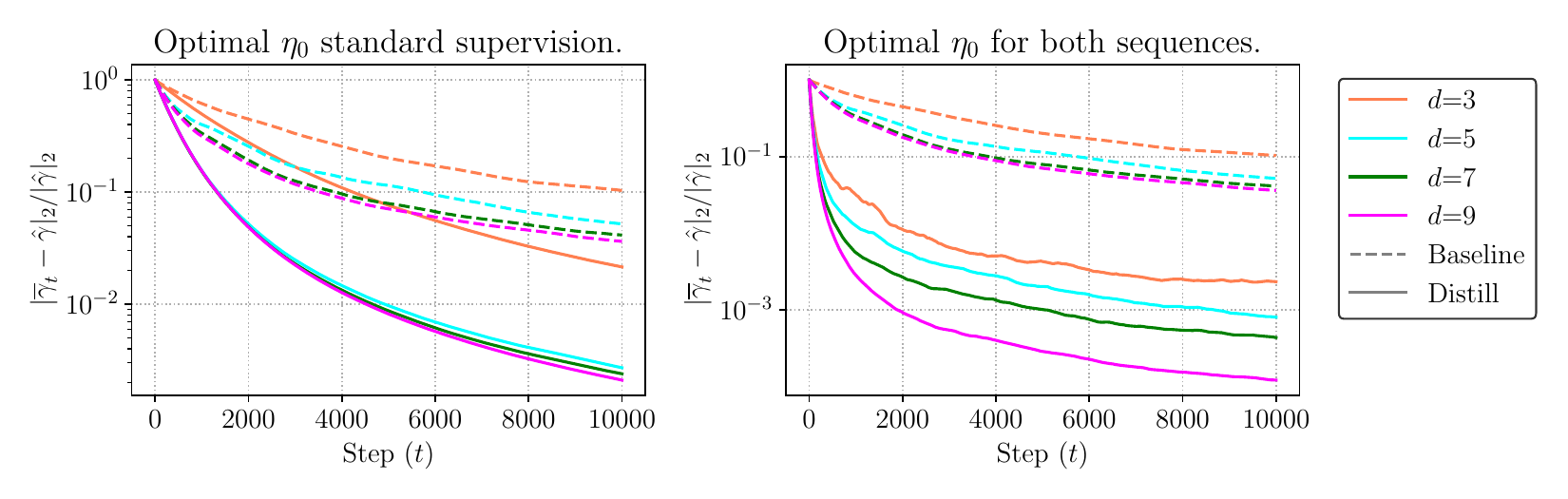}
    \caption{(Logistic regression, synthetic data) Distance from optimal point $\hat{\gamma}$ for stochastic gradient descent iterates under knowledge distillation and standard supervision. All results are averaged over 100 trials. On the left, both trajectories use $\eta_0$ that is optimal for standard supervision. On the right, the step $\eta_0$ is chosen optimally for both methods.}
    \label{fig:glm-synthetic-both}
\end{figure}

Table~\ref{tab:glm-synthetic-steps} contains the initial step size that minimizes parameter error after 10,000 steps of SGD for both knowledge distillation and standard supervision. Figure~\ref{fig:glm-synthetic-both} plots the per-step parameter error for both methods, with the experimental setup previously described in Section~\ref{subsection:synthetic-experiments}.

\begin{table}[h!]
    \centering
    \begin{tabular}{@{}rcccl@{}}
    \toprule
    \multicolumn{5}{c}{\textbf{Step size $\eta_0$, logistic regression on synthetic data.}}                              \\ \midrule
    \multicolumn{1}{l}{}          & $d=3$       & $d=5$ & $d=7$ & $d=9$ \\ \midrule
    Standard supervision $\eta_0$ & $10^{-1.5}$ & 0.1   & 0.1   & 0.1   \\
    Distillation $\eta_0$         & 1.0         & 1.0   & 1.0   & 1.0   \\ \bottomrule
    \end{tabular}
    \caption{Initial step size $\eta_0$ for logistic regression on synthetic data.}
    \label{tab:glm-synthetic-steps}
\end{table}

\subsection{Rank-1 matrix approximation}\label{appendix:rank1-synthetic}
\begin{figure}[h!]
    \centering
    \includegraphics[width=1.0\linewidth]{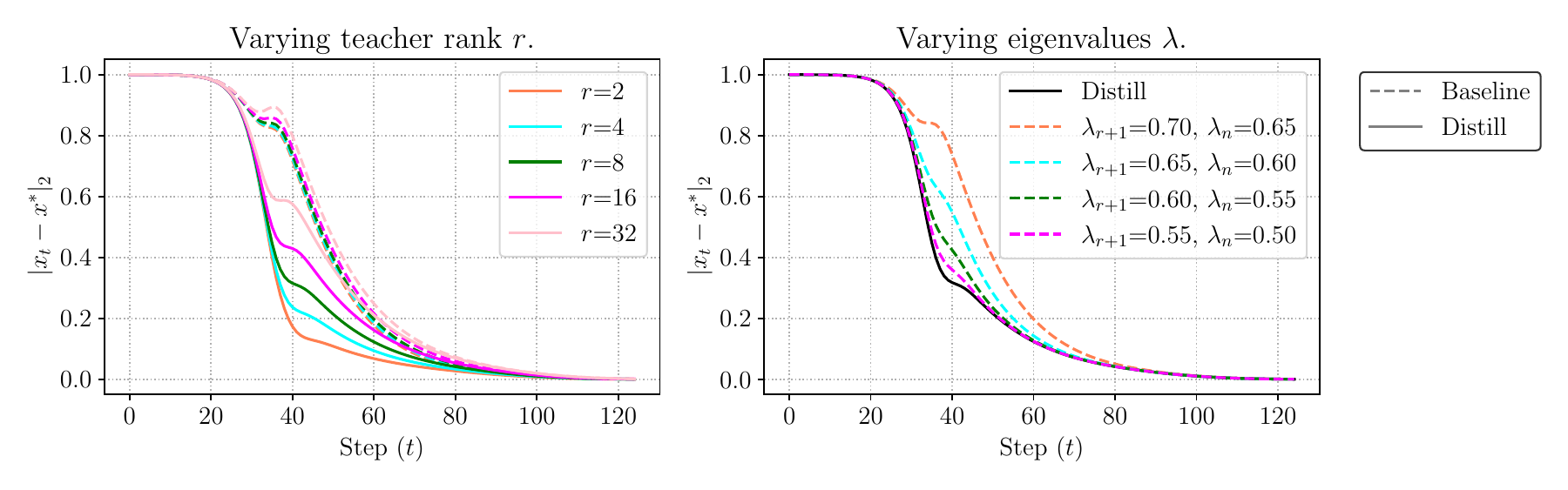}
    \caption{(Rank-1 approximation) Parameter error of gradient descent iterates under knowledge distillation and supervised learning, averaged over 100 trials. In the plot on the left, the eigenvalue spectrum is fixed ($\lambda_r = 0.7$, $\lambda_n = 0.65$) and teacher ranks are varied. In the plot on the right, the teacher rank is fixed ($r=8$) and the ``tail`` spectrum $\lambda_r$ through $\lambda_n$ is varied.}
    \label{fig:rank-1-both}
\end{figure}
We construct the target matrix with top eigenvalue $\lambda_1=1$, and the arithmetic sequence $s_r$ consists of equally spaced values from $\lambda_2=0.8$ to $\lambda_r=0.75$. Across all teacher ranks $r$ and eigenvalues $\lambda$, the step size that achieves the lowest parameter error is $\eta = 10^{-0.5}$. Figure~\ref{fig:rank-1-both} plots the error for gradient descent iterates under standard supervision and knowledge distillation.

\subsection{Multinomial logistic regression on Fashion-MNIST}\label{appendix:fmnist-best-lr}
\begin{figure}[h!]
    \centering
    \includegraphics[width=0.6\linewidth]{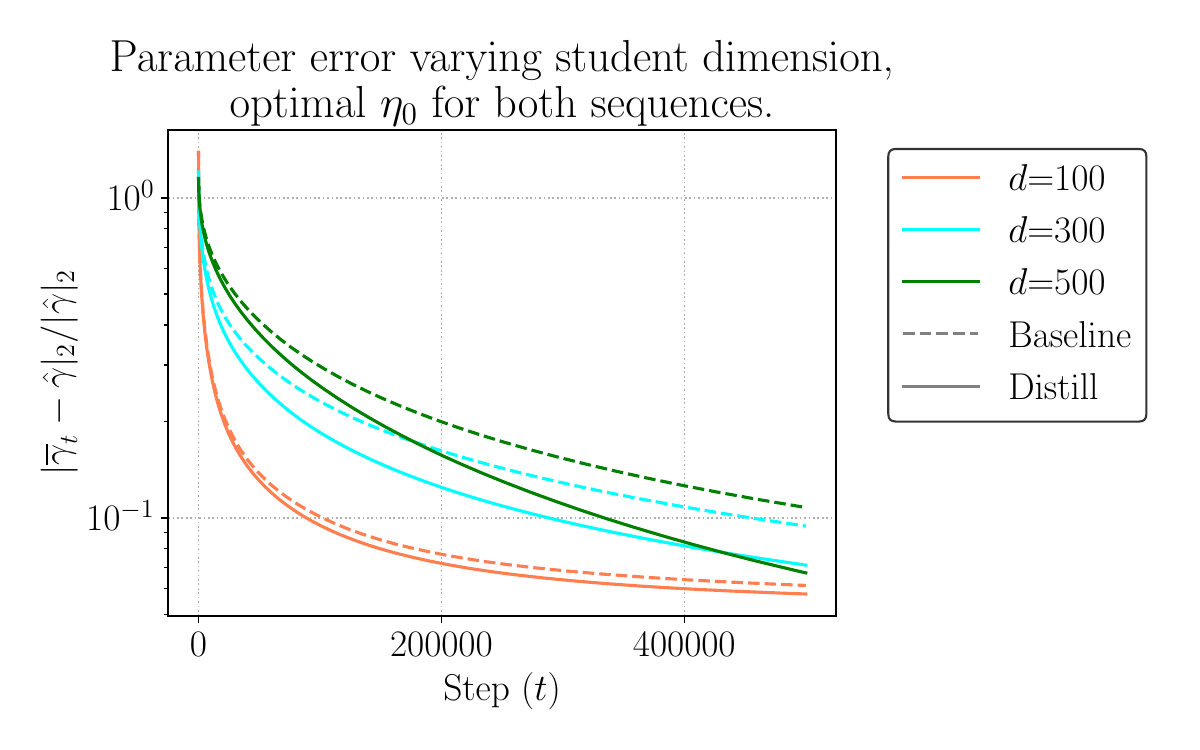}
    \caption{(Multinomial logistic regression, Fashion-MNIST) Normalized parameter error, averaged over 10 trials, of SGD iterates under knowledge distillation and standard supervision for student dimensions $d \in\{100, 300, 500\}$. The initial step size $\eta_0$ is chosen optimally for both methods.}
    \label{fig:fmnist-best-lr}
\end{figure}

We report the initial step size $\eta_0$ that minimizes parameter error after 500,000 steps of SGD for both knowledge distillation and standard supervision (Table~\ref{tab:fmnist-step}). Using these optimal step sizes, we plot the per-step parameter error for both methods (Figure~\ref{fig:fmnist-best-lr}). The optimal $\eta_0$ is the same for both methods when the student dimension is $d \in \{100, 300\}$, however, when $d=500$, distillation achieves a lower parameter error with a larger initial step size.

\begin{table}[h!]
    \centering
    \begin{tabular}{@{}rccc@{}}
        \toprule
        \multicolumn{4}{c}{\textbf{Step size $\eta_0$, multinomial logistic regression.}}                                    \\ \midrule
        \multicolumn{1}{l}{}          & $d=100$      & $d=300$     & $d=500$      \\ \midrule
        Standard supervision $\eta_0$ & $10^{-0.75}$ & $10^{-0.5}$ & $10^{-0.5}$  \\
        Distillation $\eta_0$         & $10^{-0.75}$ & $10^{-0.5}$ & $10^{-0.25}$ \\ \bottomrule
    \end{tabular}
    \caption{Initial steps size $\eta_0$ for multinomial logistic regression on the Fashion-MNIST dataset.}
    \label{tab:fmnist-step}
\end{table}

\subsection{Experiment details: Language model pretraining}\label{appendix:pretraining-details}
We draw 87.3 billion tokens from the C4 dataset as our initial pool. To estimate entropy, we train a GPT-2 medium model on 7.1 billion tokens from the dataset and compute the per-sample entropy as the average token-level entropy from its predicted distributions. We use a learning rate of $\eta = 5\cdot 10^{-4}$ to train the teacher. Based on these estimates, we partition the remaining data into 13 subsets of increasing entropy, each with a train/validation split of 6.1 billion and 30 million tokens, respectively. For each subset, we perform compute-optimal training of the student model \citep{HBMB22}, specifically a GPT-2 small model trained on 2.48 billion tokens. We sweep learning rates $\eta \in \{10^{-4}, 10^{-3.5}, 10^{-3}, 10^{-2.5}, 10^{-2}\}$ and select the model with the lowest final validation loss as our supervised learning baseline. We find that the optimal learning rate for all subsets is $\eta = 10^{-3}$. 

The distillation labels are probabilities $P_T$ from a GPT-2 medium model, trained on the full train split. The distillation student model is trained on the same 2.48 billion tokens, with the same learning rate $\eta$ and training setup as the baseline, but has the KL-divergence $D_\mathrm{KL}(P_T \ || \ P_S)$ as the objective function where $P_S$ is the student's predicted distribution. All models are trained with a batch size of 512, weight decay of 0.01, gradient clipping of 1.0, learning rate warmup ratio of 0.01, and cosine learning rate schedule. All experiments were conducted on a univeristy-managed computing cluster, using either 4x NVIDIA A100 GPUs (80GB) or 4x NVIDIA RTX 3090 (24GB).

\end{document}